\RequirePackage{fix-cm}
\documentclass[smallextended]{svjour3}       % onecolumn (second format)
\smartqed  % flush right qed marks, e.g. at end of proof
\usepackage{graphicx}
\usepackage{subfig}
\usepackage{booktabs}
\usepackage{color}
\usepackage{amssymb}
\usepackage{amsmath}
\usepackage{hyperref}
\usepackage{multirow}
\usepackage[ruled,vlined]{algorithm2e}
\usepackage{pdflscape} %Gives us landscape pages with begin{landscape}. Use this for pages with very large diagrams.
\usepackage{geometry} %change margin on individual page \newgeometry{} and \restoreometry
\usepackage{float} %exact placement of floats (things inside a begin{})
\usepackage[numbers,sort&compress]{natbib}
\usepackage{longtable}
\usepackage{mathtools}
\usepackage{csquotes}

\newcommand{\figscale}{.58}
\DeclareMathOperator*{\argmin}{arg\,min}

\begin{document}

\title{One Deep Music Representation to Rule Them All?
  \thanks{This work has been accepted to ``Neural Computing and Applications: Special Issue on Deep Learning for Music and Audio''.}
} 
%Effects of Different Learning and Evaluation \textcolor{red}{task}s}
\subtitle{A comparative analysis of different representation learning strategies}


\author{Jaehun Kim         \and
        Juli\'{a}n Urbano \and\\
        Cynthia C.~S. Liem \and 
        Alan Hanjalic%etc.
}


\institute{J. Kim, J. Urbano, C. C. S. Liem \& A. Hanjalic \at
              %Mekelweg 4 2628 CD \\
              Multimedia Computing Group \\
              Department of Intelligent Systems\\
              Faculty of Electrical Engineering, Mathematics and Computer Science\\
              Delft University of Technology \\
              Tel.: +31-15-27-87241\\
              \email{J.H.Kim@tudelft.nl}           %  \\
}

\date{Received: date / Accepted: date}

\maketitle

\begin{abstract}
Inspired by the success of deploying deep learning in the fields of Computer Vision and Natural Language Processing, this learning paradigm has also found its way into the field of Music Information Retrieval. In order to benefit from deep learning in an effective, but also efficient manner, deep transfer learning has become a common approach. In this approach, it is possible to reuse the output of a pre-trained neural network as the basis for a new learning task. The underlying hypothesis is that if the initial and new learning tasks show commonalities and are applied to the same type of input data (e.g. music audio), the generated deep representation of the data is also informative for the new task. Since, however, most of the networks used to generate deep representations are trained using a single initial learning source, their representation is unlikely to be informative for all possible future tasks. In this paper, we present the results of our investigation of what are the most important factors to generate deep representations for the data and learning tasks in the music domain. We conducted this investigation via an extensive empirical study that involves multiple learning sources, as well as multiple deep learning architectures with varying levels of information sharing between sources, in order to learn music representations. We then validate these representations considering multiple target datasets for evaluation. The results of our experiments yield several insights on how to approach the design of methods for learning widely deployable deep data representations in the music domain.
\end{abstract}

\newpage

\section{Introduction}
\label{intro}

In the Music Information Retrieval (MIR) field, many research problems of interest involve the automatic description of properties of musical signals, employing concepts that are understood by humans. For this, tasks are derived that can be solved by automated systems. In such cases, algorithmic processes are employed to map raw music audio information to humanly understood descriptors (e.g.\ genre labels or descriptive tags). To achieve this, historically, the raw audio would first be transformed into a \emph{representation} based on \emph{hand-crafted features}, which are engineered by humans to reflect dedicated semantic signal properties. The feature representation would then serve as input to various statistical or Machine Learning (ML) approaches~\cite{Casey2008Content-basedChallenges}.\par

The framing as described above can generally be applied to many applied ML problems: complex real-world problems are abstracted into a relatively simpler form, by establishing tasks that can be computationally addressed by automatic systems. In many cases, the task involves making a prediction based on a certain observation. For this, modern ML methodologies can be employed, that automatically can infer the logic for the prediction directly from (a numeric representation of) the given data, by optimizing an objective function defined for the given task.

However, music is a multimodal phenomenon, that can be described in many parallel ways, ranging from objective descriptors to subjective preference.
As a consequence, in many cases, while music-related tasks are well understood by humans, it often is hard to pinpoint and describe where the truly `relevant' information is in the music data used for the tasks, and how this properly can be translated into numeric representations that should be used for prediction. While research into such proper translations can be conducted per individual task, it is likely that informative factors in music data will be shared across tasks. As a consequence, when seeking to identify informative factors that are not explicitly restricted to a single task, Multi-Task Learning (MTL) is a promising strategy. In MTL, a single learning framework hosts multiple tasks at once, allowing for models to perform better by sharing commonalities between involved tasks~\cite{RichCaruana1997Multitask}. MTL has been successfully used in a range of applied ML works~\cite{Bengio2012RepresentationPerspectives,Liu2015Multi-taskSelection, Bingel2017IdentifyingNetworks,li2014heterogeneous,Zhang2015DeepAnalysis,ZhangFacialLearning, DBLP:journals/corr/KaiserGSVPJU17,DBLP:conf/iccv/ChangLPK17}, also including the music domain~\cite{Weston2011Multi-TaskingRetrieval,Aytar2016SoundNet:Video}.

Following successes in the fields of Computer Vision (CV) and Natural Language Processing (NLP), deep learning approaches have recently also gained increasing interest in the MIR field, in which case \emph{deep representations} of music audio data are directly learned from the data, rather than being hand-crafted. Many works employing such approaches reported considerable performance improvements in various music analysis, indexing and classification tasks~\cite{Hamel2010LearningNetworks,Boulanger-Lewandowski2012ModelingTranscription,schlueter2014_icassp,Choi2016AutomaticNetworks,Oord2013DeepRecommendation,chandna2017monoaural,Jeong2016LearningClassification,Han2016DeepMusic}.\par

In many deep learning applications, rather than training a complete network from scratch, pre-trained networks are commonly used to generate deep representations, which can be either directly adopted or further adapted for the current task at hand. In CV and NLP, (parts of) certain pre-trained networks ~\cite{Simonyan2014VeryRecognition,he2016deep,Szegedy2015GoingConvolutions,Mikolov2013EfficientSpace} have now been adopted and adapted in a very large number of works. These `standard' deep representations have typically been obtained by training a network for a single learning task, such as visual object recognition, employing large amounts of training data. The hypothesis on why these representations are effective in a broader of spectrum of tasks than they originally were trained for, is that \emph{deep transfer learning (DTL)} is happening: information initially picked up by the network is beneficial also for new learning tasks performed on the same type of raw input data. Clearly, the validity of this hypothesis is linked to the extent to which the new task can rely on similar data characteristics as the task on which the pre-trained network was originally trained.\par

Although a number of works deployed DTL for various learning tasks in the music domain\cite{Dieleman2011Audio-basedNetwork,Choi2017TransferTasks,van2014transfer,Liang2014Content-AwareNetworks}, to our knowledge, however, transfer learning and the employment of pre-trained networks are not as standard in the MIR domain as in the CV domain. Again, this may be due to the broad and partially subjective range and nature of possible music descriptions. Following the considerations above, it may then be useful to combine deep transfer learning with multi-task learning.

\begin{figure}
\centering
\includegraphics[height=0.3\textheight]{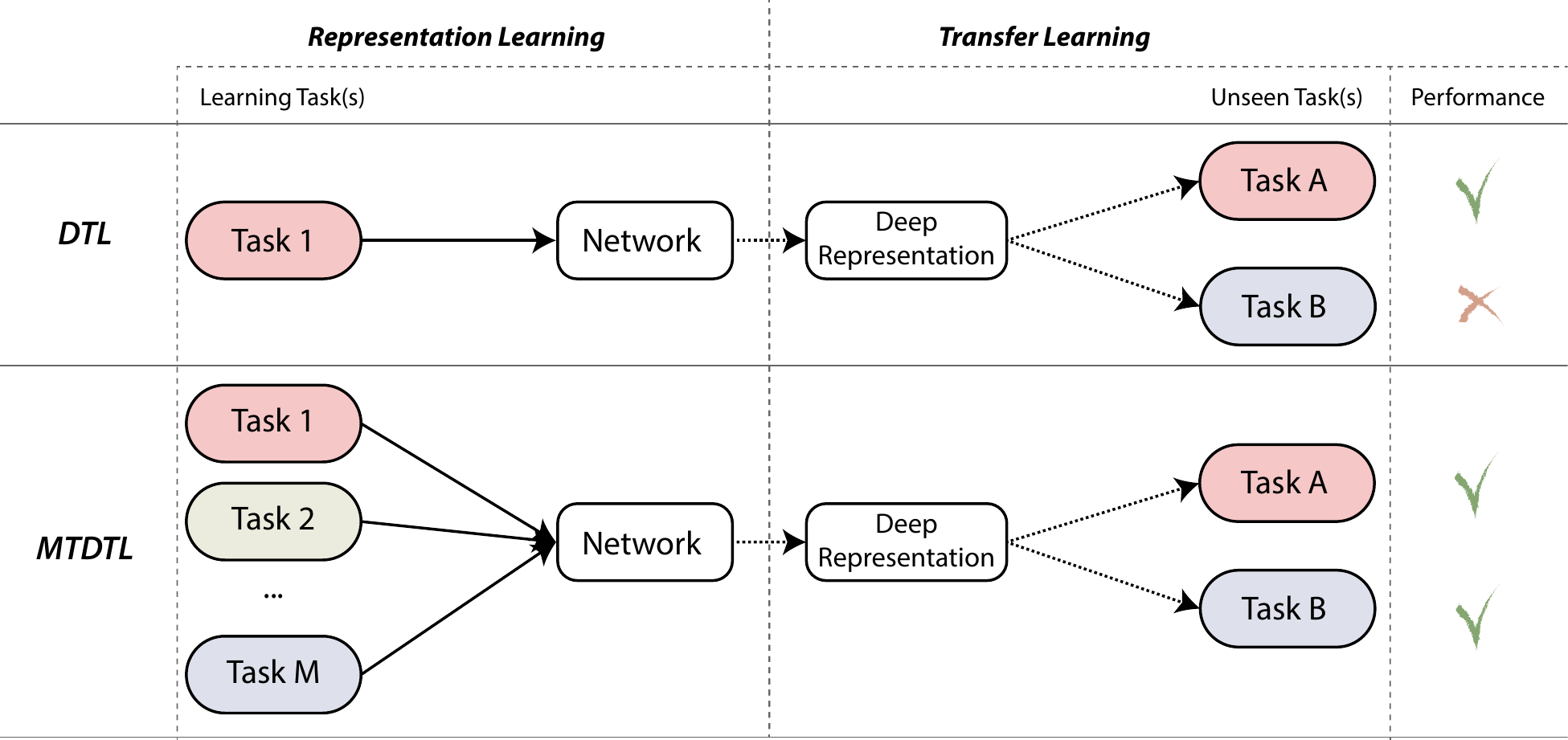}
\caption{Simplified illustration of the conceptual difference between traditional deep transfer learning (DTL) based on a single learning task (above) and multi-task based deep transfer learning (MTDTL) (below). The same color used for a learning and an target task indicates that the tasks have commonalities, which implies that the learned representation is likely to be informative for the target task. At the same time, this representation may not be that informative to another future task, leading to a low transfer learning performance. The hypothesis behind MTDTL is that relying on more learning tasks increases robustness of the learned representation and its usability for a broader set of target tasks.}
\label{fig:toyexample}
\end{figure}

Indeed, in order to increase robustness to a larger scope of new learning tasks and datasets, the concept of MTL also has been applied in training deep networks for representation learning, both in the music domain ~\cite{Aytar2016SoundNet:Video,Weston2011Multi-TaskingRetrieval} and in general~\cite[p.~2]{Bengio2012RepresentationPerspectives}. As the model learns several tasks and datasets in parallel, it may pick up commonalities among them. As a consequence, the expectation is that a network learned with MTL will yield robust performance across different tasks, by transferring shared knowledge~\cite{RichCaruana1997Multitask,Bengio2012RepresentationPerspectives}. A simple illustration of the conceptual difference between traditional DTL and deep transfer learning based on MTL (further referred to as \emph{multi-task based deep transfer learning (MTDTL))} is shown in Fig.~\ref{fig:toyexample}.\par

The mission of this paper is to investigate the effect of conditions around the setup of MTDTL, which are important to yield effective deep music representations. Here, we understand an `effective' representation to be a representation that is suitable for a wide range of new tasks and datasets. 
Ultimately, we aim for providing a methodological framework to systematically obtain and evaluate such transferable representations. We pursue this mission by exploring the effectiveness of MTDTL and traditional DTL, as well as concatenations of multiple deep representations, obtained by networks that were independently trained on separate single learning tasks. We consider these representations for multiple choices of learning tasks and considering multiple target datasets.

Our work will address the following research questions:

\begin{itemize}
\item \textbf{RQ1:} Given a set of learning sources that can be used to train a network, what is the influence of the number and type of the sources on the effectiveness of the learned deep representation? 
\item \textbf{RQ2:} How do various degrees of information sharing in the deep architecture affect the effectiveness of a learned deep representation?
\end{itemize}

By answering the \textbf{RQ1} we arrive at an understanding of important factors regarding the composition of a set of learning tasks and datasets (which in the remainder of this work will be denoted as \emph{learning sources}) to achieve an effective deep music representation, specifically on the number and nature of learning sources. The answer to \textbf{RQ2} provides insight in \emph{how to choose the optimal multi-task network architecture} under a MTDTL context. For example, in MTL, multiple sources are considered under a joint learning scheme, that partially shares inferences obtained from different learning sources in the learning pipeline. In MTL applications using deep neural networks, this means that certain layers will be shared between all sources, while at other stages, the architecture will `branch' out into source-specific layers~\cite{RichCaruana1997Multitask,Bingel2017IdentifyingNetworks,li2014heterogeneous,Zhang2015DeepAnalysis,ZhangFacialLearning,misra2016cross,Aytar2016SoundNet:Video}. However, investigation is still needed on where in the layered architecture branching should ideally happen---if a branching strategy would turn out beneficial in the first place.\par

To reach the aforementioned answers, it is necessary to conduct a systematic assessment to examine relevant factors. For \textbf{RQ1}, we investigate different numbers and combinations of learning sources. For \textbf{RQ2}, we study different architectural strategies. However, we wish to ultimately investigate effectiveness of the representation with respect to new, target learning tasks and datasets (which in the remainder of this paper will be denoted by \emph{target datasets}). While this may cause combinatorial explosion with respect to possible experimental configurations, we will make strategic choices in the design and evaluation procedure of the various representation learning strategies.\par

The scientific contribution of this work can be summarized as follows:

\begin{itemize}
  \item[$\bullet$] We provide insight into the effectiveness of various deep representation learning strategies under the multi-task learning context. 
  \item[$\bullet$] We offer in-depth insight into ways to evaluate desired properties of a deep representation learning procedure.
  \item[$\bullet$] We propose and release several pre-trained music representation networks, based on different learning strategies for multiple semantic learning sources.
\end{itemize}

The rest of this work is presented as following: a formalization of this problem, as well as the global outline of how learning will be performed based on different learning tasks from different sources, will be presented in Section~\ref{learning_framework}. Detailed specifications of the deep architectures we considered for the learning procedure will be discussed in Section~\ref{dl_specifications}. Our strategy to \emph{evaluate} the effectiveness of different representation network variants by employing various \emph{target datasets} will be the focus of Section~\ref{eval}. Experimental results will be discussed in Section~\ref{res:intro}, after which general conclusions will be presented in Section~\ref{concl}.

\section{Framework for Deep Representation Learning}
\label{learning_framework}

\begin{figure}[!htp]
    \centering
    \subfloat[Multi-Task Transfer Learning in General Problem Domain]{%
        \includegraphics[width=0.7\textwidth]{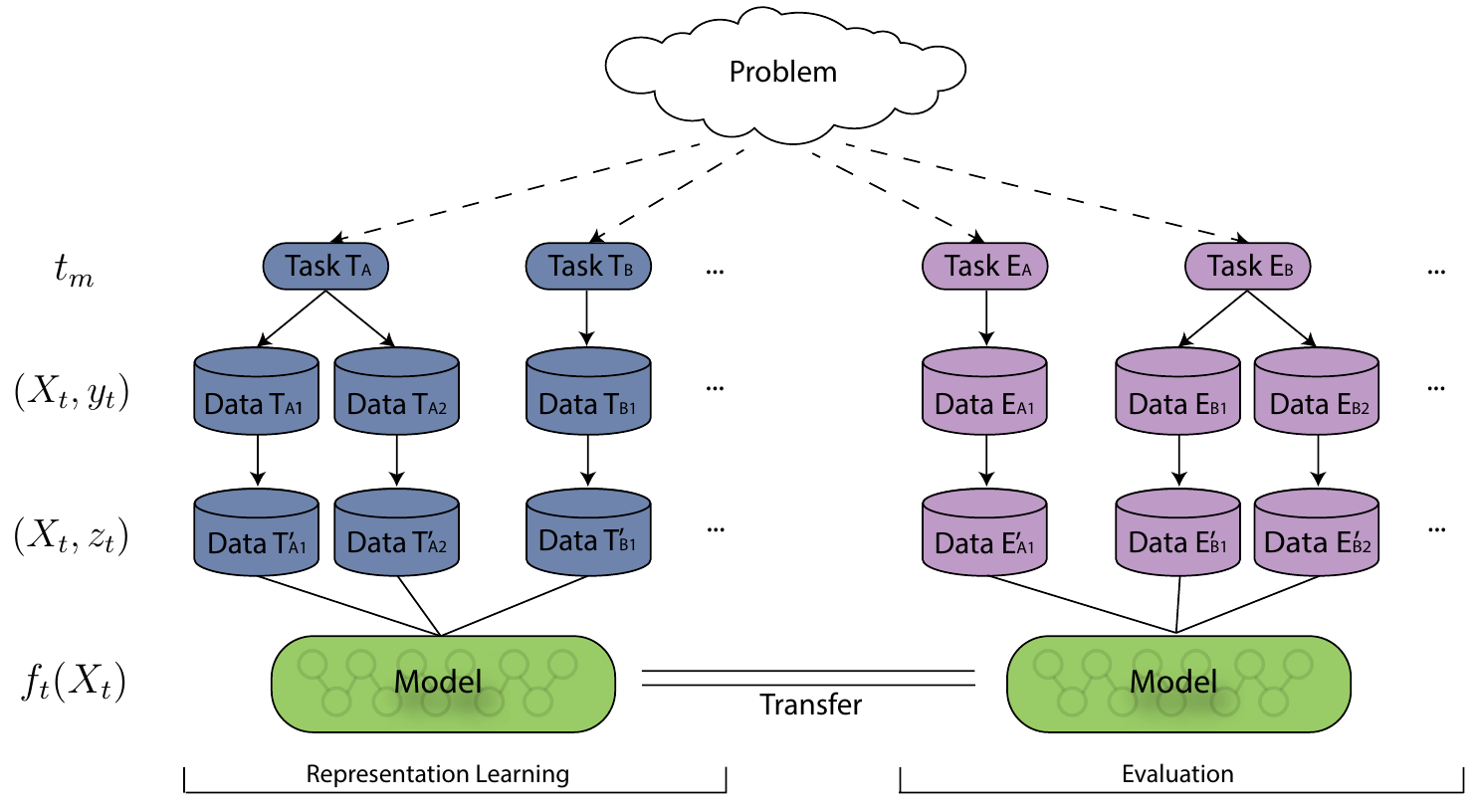}%
        \label{fig:problem:general}%
        }%
    \hfill%
        \subfloat[Multi-Task Transfer Learning in Music Information Retrieval Domain]{%
        \includegraphics[width=0.7\textwidth]{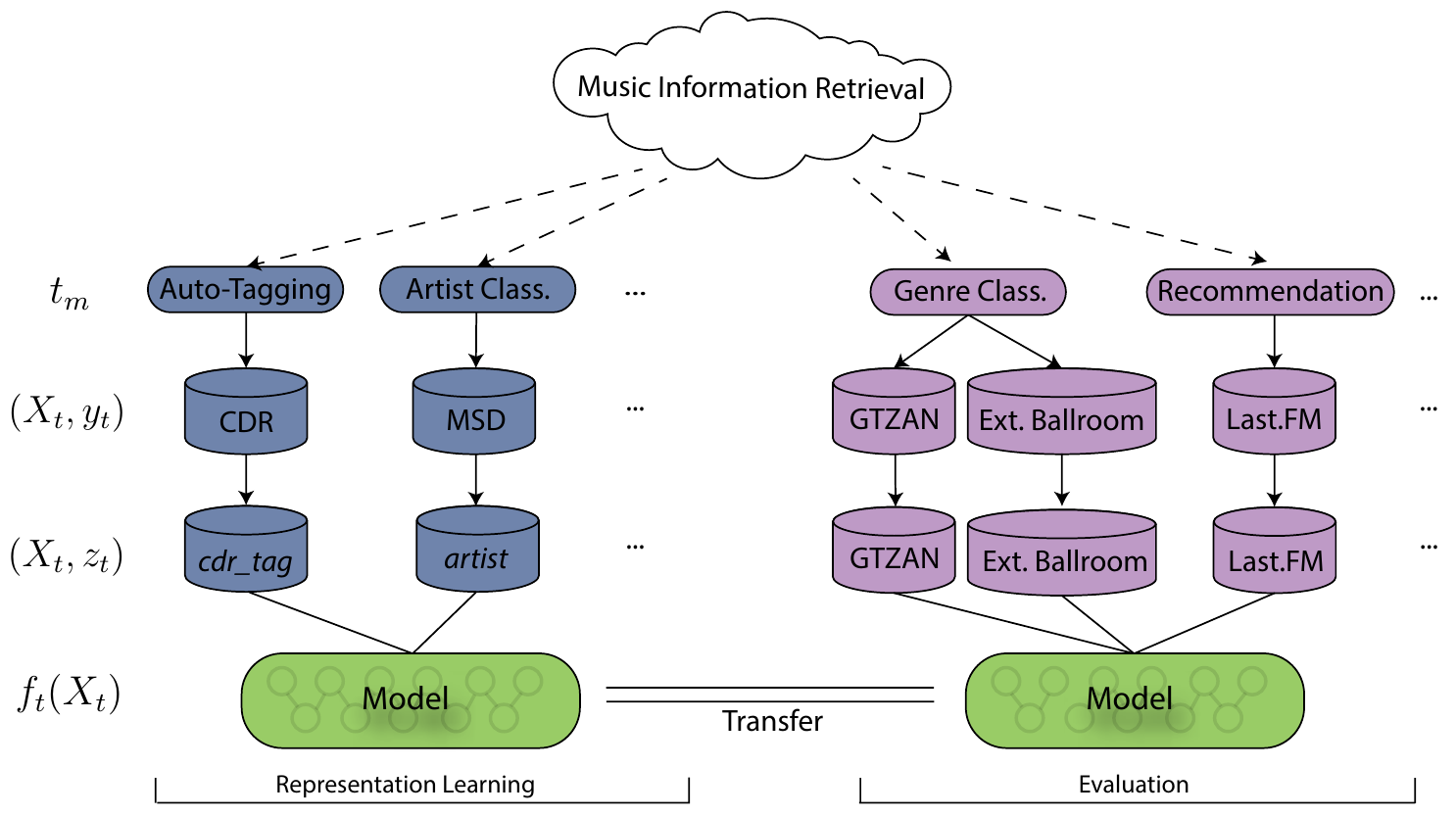}%
        \label{fig:problem:example}%
        }%
    \hfill%
    \caption{Schematic overview of what this work investigates. The upper scheme illustrates a general problem solving framework in which multi-task transfer learning is employed. The tasks $t \in \{t_0, t_1, \cdots, t_M\}$ are derived from a certain problem domain, which are instantiated by datasets, that often are represented as sample pairs of observations and corresponding labels $(X_{t}, y_{t})$. Sometimes, the original dataset is processed further into simpler representation forms $(X_{t}, z_{t})$, to filter out undesirable information and noise. Once a model or system $f_{t}(X_{t})$ has learned the necessary mappings within the learning sources, this knowledge can be transferred to another set of target datasets, leveraging commonalities already obtained by the pre-training. Below the general framework, we show a concrete example, in which the broad MIR problem domain is abstracted into various sub-problems with corresponding tasks and datasets.}
    \label{fig:problem}
\end{figure}

In this section, we formally define the deep representation learning problem. As Fig.~\ref{fig:problem} illustrates, any domain-specific MTDTL problem can be abstracted into a formal task, which is instantiated by a specific dataset with specific observations and labels. Multiple tasks and datasets are involved to emphasize different aspects of the input data, such that the learned representation is more adaptable to different future tasks. The learning part of this scheme can be understood as the MTL phase, which is introduced in Section~\ref{learning_framework:prob_def}. Subsequently in Section~\ref{learning_framework:learning_sources}, we discuss learning sources involved in this work, which consist of various tasks and datasets to allow investigating their effects on the transfer learning. Further, we introduce the label preprocessing procedure that is applied in this work in Section~\ref{learning_framework:learning_sources:learnfromfactors}, ensuring that the learning sources are more regularized, such that their comparative analysis is clearer.

\subsection{Problem Definition}
\label{learning_framework:prob_def}

A machine learning problem, focused on solving a specific task $t$, can be formulated as a minimization problem, in which a model function $f_t$ must be learned that minimizes a loss function $\mathcal{L}$ for given dataset $\mathcal{D}_{t} =  \{\,(x^{(i)}_{t}, y^{(i)}_{t}) \mid i \in \{1, \cdots, I\} \,\}$, comparing the model's predictions given by the input $x_t$ and actual task-specific learning labels $y_t$. This can be formulated using the following expression:

\begin{equation} \label{eq:1}
\hat\theta = \argmin\;\mathbb{E}_{\mathcal{D}_{t}}\mathcal{L}(y_t, f_t(x_t;\theta))
\end{equation}

where $x_{t}\in\mathbb{R}^d$ is, traditionally, a hand-crafted $d$-dimensional feature vector and $\theta$ is a set of model parameters of $f$.

When deep learning is employed, the model function $f$ denotes a learnable network. Typically, the network model $f$ is learned in an end-to-end fashion, from raw data at the input to the learning label. In the speech and music field, however, using true end-to-end learning is still not a common practice. Instead, raw data is typically transformed first, before serving as network input. More specifically, in the music domain, common input to function $f$ would be $X\in\mathbb{R}^{c\times{n}\times{b}}$, replacing the originally hand-crafted feature vector $x\in\mathbb{R}^d$ from (\ref{eq:1}) by a time-frequency representation of the observed music data, usually obtained through the Short-Time Fourier Transform (STFT), with potential additional filter bank applications (e.g.\ mel-filter bank). The dimensions $c$, $n$, $b$ indicate channels of the audio signal, time steps, and frequency bins respectively.

If such a network still is trained for a specific single machine learning task $t$, we can now reformulate (\ref{eq:1}) as follows:

\begin{equation} \label{eq:2}
\hat\theta = \argmin \; \mathbb{E}_{\mathcal{D}_{t}}\mathcal{L}(y_{t}, f_{t}(X_{t};\theta)).
\end{equation}

In MTL, in the process of learning the network model $f$, different tasks will need to be solved in parallel. In case of deep neural networks, this is usually realized by having a network in which lower layers are shared for all tasks, but upper layers are task-specific. Given $m$ different tasks $t$, each having the learning label $y_{t}$, we can formulate the learning objective of the neural network in a MTL scenario as follows:

\begin{equation} \label{eq:4}
\hat\theta^{s}, \hat\theta^{*} = \argmin \; \mathbb{E}_{t\in{\mathcal{T}}}\mathbb{E}_{\mathcal{D}_{t}}\mathcal{L}(y_{t}, f_{t}(X_{t};\theta^{s},\theta^{t}))
\end{equation}

Here, $\mathcal{T}=\{t_{1},t_{2},...,t_{m}\}$ is a given set of tasks to be learned and $\theta^{*}=\{\theta^{1},\theta^{2},...,\theta^{m}\}$ indicates a set of model parameters $\theta^{t}$ with respect to each task. Since the deep architecture initially shares lower layers and branches out to task-specific upper layers, the parameters of shared layers and task-specific layers are referred to separately as $\theta^{s}$ and $\theta^{t}$, respectively. Updates for all parameters can be achieved through standard back-propagation. Further specifics on network architectures and training configurations will be given in Section~\ref{dl_specifications}.\par
Given the formalizations above, the first step in our framework is to select a suitable set $\mathcal{T}$ of learning tasks. These tasks can be seen as multiple concurrent descriptions or transformations of the same input fragment of musical audio: each will reflect certain semantic aspects of the music.
However, unlike the approach in a typical MTL scheme, solving multiple specific learning tasks is actually not our main goal; instead, we wish to learn an effective \emph{representation} that captures as many semantically important factors in the low-level music representation as possible. Thus, rather than using learning labels $y_{t}$, our representation learning process will employ reduced learning labels $z_{t}$, which capture a reduced set of semantic factors from $y_{t}$. We then can reformulate (\ref{eq:4}) as follows:

\begin{equation} \label{eq:5}
\hat\theta^{s}, \hat\theta^{*} = \argmin \; \mathbb{E}_{t\in{\mathcal{T}}}\mathbb{E}_{\mathcal{D}_{t}}\mathcal{L}(z_{t}, f_{t}(X_{t};\theta^{s},\theta^{t}))
\end{equation}

where $z_t\in\mathbb{R}^{k}$ is a $k$-dimensional vector that represents reduced learning label for a specific task $t$. Each $z_t$ will be obtained through task-specific factor extraction methods, as described in Section~\ref{learning_framework:learning_sources:learnfromfactors}.

\subsection{Learning Sources}
\label{learning_framework:learning_sources}

In the MTDTL context, a training dataset can be seen as the `source' to learn the representation, which will be further transferred to the future `target' dataset. Different learning sources of different nature can be imagined, that can be globally categorized as \emph{Algorithm} or \emph{Annotation}. As for the \emph{Algorithm} category, by employing traditional feature extraction or representation transformation algorithms, we will be able to automatically extract semantically interesting aspects from input data. As for the \emph{Annotation} category, these include different types of label annotations of the input data by humans.\par

The dataset used as resource for our learning experiments is the Million Song Dataset (MSD)\cite{Bertin-Mahieux2011}. In its original form, it contains metadata and precomputed features for a million songs, with several associated data resources, e.g.\ considering \texttt{Last.fm} social tags and listening profiles from \texttt{the Echo Nest}. While the MSD does not distribute audio due to copyright reasons, through the API of the \texttt{7digital} service, 30-second audio previews can be obtained for the songs in the dataset. These 30-second previews will form the source for our raw audio input.\par

Using the MSD data, we consider several subcategories of learning sources within the \emph{Algorithm} and \emph{Annotation} categories; below, we give an overview of these, and specify what information we considered exactly for the learning labels in our work.

\subsubsection{Algorithm}
\label{learning_framework:learning_sources:algorithm}

\begin{itemize}
\item \textbf{\textit{Self.}} The music track is the learning source itself; in other words, intrinsic information in the input music track should be captured through a learning procedure, without employing further data. Various unsupervised or auto-regressive learning strategies can be employed under this category, with variants of Autoencoders, including the Stacked Autoencoder~\cite{bengio2007greedy,Vincent2008ExtractingAutoencoders}, Restricted Boltzmann Machines (RBM)~\cite{smolensky1986information}, Deep Belief Networks (DBN)~\cite{Hinton2006ANets} and Generative Adversarial Networks (GAN)~\cite{goodfellow2014generative}. As another example within this category, variants of the Siamese networks for similarity learning can be considered~\cite{Han2015MatchNet:Matching,Arandjelovic2017LookLearn,Huang2017SimilarityGames}.

In our case, we will employ the Siamese architecture to learn a metric that measures whether two input music clips belong to the same track, or two different tracks. This can be formulated as follows:

\begin{equation} \label{eq:self}
\hat\theta^{self}, \hat\theta^{s} = \argmin \; \mathbb{E}_{X_l, X_r \sim \mathcal{D}_{self}} \mathcal{L}(y_{self}, f_{self}(X_{l},X_{r};\theta^{self},\theta^{s}))
\end{equation}

\begin{equation} \label{eq:self_h}
y_{self}= 
\begin{cases}
    1, & \text{if } X_{l} \text{ and } X_{r} \text{ sampled from same track} \\
    0 & \text{otherwise}
\end{cases}
\end{equation}

where $X_{l}$ and $X_{r}$ are a pair of randomly sampled short music snippets (taken from the 30-second MSD audio previews) and $f_{self}$ is a network for learning a metric between given input representations in terms of the criteria imposed by $y_{self}$. It is composed of one or more fully-connected layers and one output layer with softmax activation. An global outline illustration of our chosen architecture is given in Fig.~\ref{fig:match_arch}. Further specifications of the representation network and sampling strategies will be given in Section~\ref{fig:match_arch}.

\begin{figure}[htp]
    \centering
	\includegraphics[height=0.33\textheight]{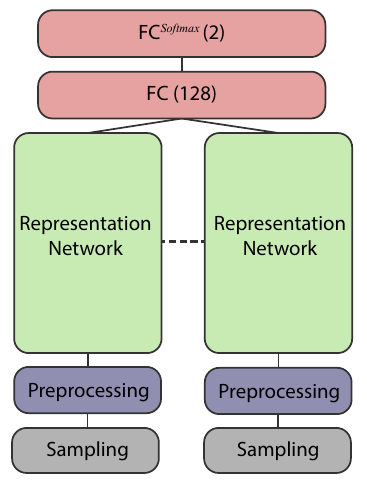}
    \caption{Siamese architecture adopted for the \emph{self} learning task. For further details of the Representation Network, see Section~\ref{dl_specifications:base_architecture} and Fig.~\ref{fig:base_arch}.}
    \label{fig:match_arch}
\end{figure}

\item \textbf{\textit{Feature.}} Many algorithms exist already for extracting features out of musical audio, or for transforming musical audio representations. By running such algorithms on musical audio, learning labels are automatically computed, without the need for soliciting human annotations.
Algorithmically computed outcomes will likely not be perfect, and include noise or errors. At the same time, we consider them as a relatively efficient way to extract semantically relevant and more structured information out of a raw input signal.\par
In our case, under this category, we use Beat Per Minute (BPM) information, released as part of the MSD's precomputed features. The BPM values were computed by an estimation algorithm, as part of the \texttt{Echo Nest} API.\par
\end{itemize}

\subsubsection{Annotation}
\label{learning_framework:learning_sources:annotation}

\begin{itemize}
\item \textbf{\textit{Metadata.}} Typically, metadata will come `for free' with music audio, specifying side information, such as a release year, the song title, the name of the artist, the corresponding album name, and the corresponding album cover image. Considering that this information describes categorization facets of the musical audio, metadata can be a useful information source to learn a music representation. In our experiments, we use release year information, which is readily provided as metadata with each song in the MSD.\par

\item \textbf{\textit{Crowd.}} Through interaction with music streaming or scrobbling services, large numbers of users, also designated as the \textit{crowd}, left explicit or implicit information regarding their perspectives on musical content. For example, they may have created social tags, ratings, or social media mentionings of songs. With many services offering API access to these types of descriptors, crowd data therefore offers scalable, spontaneous and diverse (albeit noisy) human perspectives on music signals.\par
In our experiments, we use social tags from \texttt{Last.fm}\footnote{\url{https://labrosa.ee.columbia.edu/millionsong/lastfm}} and user listening profiles from the \texttt{Echo Nest}.

\item \textbf{\textit{Professional.}} As mentioned in \cite{Casey2008Content-basedChallenges}, annotation of music tracks is a complicated and time-consuming process: annotation criteria frequently are subjective, and considerable domain knowledge and annotation experience may be required before accurate and consistent annotations can be made. Professional experts in categorization have this experience, and thus are capable of indicating clean and systematic information about musical content. It is not trivial to get such professional annotations at scale; however, these types of annotations may be available in existing professional libraries.\par
In our case, we use professional annotations from the Centrale Discotheek Rotterdam (CDR), the largest music library in The Netherlands, holding all music ever released in the country in physical and digital form in its collection. The CDR collection can be digitally accessed through the online Muziekweb\footnote{\url{https://www.muziekweb.nl/}} platform. For each musical album in the CDR collection, genre annotations were made by a professional annotator, according to a fixed vocabulary of 367 hierarchical music genres.\par
As another professional-level `description', we adopted lyrics information per each track, which is provided in Bag-of-Words format with the MSD. To filter out trivial terms such as stop-words, we applied TF-IDF\cite{salton1983introduction}.\par

\item \textbf{\textit{Combination.}} Finally, learning labels can be derived from combinations of the above categories. In our experiment, we used combination of artist information and social tags, by making a bag of tags at the artist level as a learning label.\par

\end{itemize}

Not all songs in the MSD actually include learning labels from all the sources mentioned above. Clearly, it is another advantage of using MTL that one can use such unbalanced datasets in a single learning procedure, to maximize the coverage of the dataset. However, on the other hand, if one uses an unbalanced number of samples across different learning sources, it is not trivial to compare the effect of individual learning sources. We therefore choose to work with a subset of the dataset, in which equal numbers of samples across learning sources can be used. As a consequence, we managed to collect 46,490 clips of tracks with corresponding learning source labels. A 41,841 / 4,649 split was made for training and validation for all sources from both MSD and CDR. Since we mainly focus on transfer learning, we used the validation set mostly for monitoring the training, to keep the network from overfitting.\par

\begin{table}
\centering
\caption{Properties of learning sources.}
\label{tab:intertask}
\begin{tabular}{llllrl}
\hline\noalign{\smallskip}
Identifier & \multicolumn{2}{c}{Category} & Data & Dimensionality & Preprocessing \\
\noalign{\smallskip}\hline\noalign{\smallskip}
\textit{self} & \multirow{ 2}{*}{Algorithm} & Self & MSD - Track & 1 & \\
\textit{bpm} & & Feature & MSD - BPM & 1 & GMM \\
\noalign{\smallskip}\hline\noalign{\smallskip}
\textit{year} & \multirow{ 6}{*}{Annotation} & Metadata & MSD - Year & 1 & GMM \\
\textit{tag} & & Crowd & MSD - Tag & 174,156 & pLSA \\
\textit{taste} & & Crowd & MSD - Taste & 949,813 & pLSA \\
\textit{cdr\_tag} & & Professional & CDR - Tag & 367 & pLSA \\
\textit{lyrics} & & Professional & MSD - Lyrics & 5,000 & pLSA, TF-IDF\\
\textit{artist} & & Combination & MSD - Artist \& Tag & 522,366 & pLSA \\
\noalign{\smallskip}\hline
\end{tabular}
\end{table}

\begin{table}
\scriptsize
\centering
\caption{Examples of Latent Topics extracted with pLSA from MSD social tags}
\label{tab:topic_term}
\begin{tabular}{ll}
\hline\noalign{\smallskip}
Topic & Strongest social tags\\
\noalign{\smallskip}\hline\noalign{\smallskip}
tag1 & \texttt{indie rock}, \texttt{indie}, \texttt{british}, \texttt{Scottish}\\
tag2 & \texttt{pop}, \texttt{pop rock}, \texttt{dance}, \texttt{male vocalists}\\
tag3 & \texttt{soul}, \texttt{rnb}, \texttt{funk}, \texttt{Neo-Soul}\\
tag4 & \texttt{Melodic Death Metal}, \texttt{black metal}, \texttt{doom metal}, \texttt{Gothic Metal}\\
tag5 & \texttt{fun}, \texttt{catchy}, \texttt{happy}, \texttt{Favorite}\\
\noalign{\smallskip}\hline
\end{tabular}
\end{table}

\subsection{Latent Factor Preprocessing}
\label{learning_framework:learning_sources:learnfromfactors}

Most learning sources are noisy. For instance, social tags include tags for personal playlist management, long sentences, or simply typos, which do not actually show relevant nuances in describing the music signal. The algorithmically extracted BPM information also is imperfect, and likely contains octave errors, in which BPM is under- or overestimated by a factor of 2. To deal with this noise, several previous works using the MSD~\cite{Choi2016AutomaticNetworks,Choi2017TransferTasks} applied a frequency-based filtering strategy along with top-down domain knowledge. However, this shrinks the available sample size. As an alternative way to handle noisiness, several other previous works~\cite{Lamere2008SocialRetrieval,Weston2011Multi-TaskingRetrieval,Hamel2013TRANSFERSIMILARITY,Law2010LearningLabels,van2014transfer,Oord2013DeepRecommendation} apply latent factor extraction using various low-rank approximation models to preprocess the label information. We also choose to do this in our experiments.

A full overview of chosen learning sources, their category, origin dataset, dimensionality and preprocessing strategies is shown in Table~\ref{tab:intertask}. In most cases, we apply probabilistic latent semantic analysis (pLSA), which extracts latent factors as a multinomial distribution of latent topics~\cite{DBLP:conf/uai/Hofmann99}. Table~\ref{tab:topic_term} illustrates several examples of strong social tags within extracted latent topics.

For situations in which learning labels are a scalar, non-binary value (BPM and release year), we applied a Gaussian Mixture Model (GMM) to transform each value into a categorical distribution of Gaussian components. In case of the \textit{Self} category, as it basically is a binary membership test, no factor extraction was needed in this case.

After preprocessing, learning source labels $y_t$ are now expressed in the form of probabilistic distributions $z_t$. Then, the learning of a deep representation can take place by minimizing the Kullback\textendash Leibler (KL) divergence between model inferences $f_t(X)$ and label factor distributions $z_t$.

Along with the noise reduction, another benefit from such preprocessing is the regularization of the scale of the objective function between different tasks involved in the learning, when the resulting factors have the same size. This regularity between the objective functions is particularly helpful for comparing different tasks and datasets. For this purpose, we used a fixed single value $k=50$ for the number of factors (pLSA) and the number of Gaussians (GMM). In the remainder of this paper, the datasets and tasks processed in above manner will be denoted by \textit{learning sources} for coherent presentation and usage of the terminology.\par

\section{Representation Network Architectures}
\label{dl_specifications}

In this section, we present the detailed specification of the deep representation neural network architecture we exploited in this work. We will discuss the base architecture of the network, and further discuss the shared architecture with respect to different fusion strategies that one can take in the MTDTL context. Also, we introduce details on the preprocessing related to the input data served into networks.\par

\subsection{Base Architecture}
\label{dl_specifications:base_architecture}

\begin{table}
\centering
\caption{Configuration of the base CNN. \texttt{conv} and \texttt{max-pool} indicate a 2-dimensional convolution and max-pooling layer, respectively. We set the stride size with 2 on the time dimension of \texttt{conv1}, to compress dimensionality at the early stage. Otherwise, all strides are set as 1 across all the convolution layers. \texttt{gap} corresponds to the global average pooling used in~\cite{he2016deep}, which averages out all the spatial dimensions of the filter responses. \texttt{fc} is an abbreviation of fully-connected layer. We use \texttt{dropout} with $p=0.5$ only for the \texttt{fc-feature} layer, where the intermediate latent representation is extracted and evaluated. For simplicity, we omit the batch-size dimension of the input shape.}
\label{tab:netarch}
\begin{tabular}{lllll}
\hline\noalign{\smallskip}
Layer & Input Shape & Weight Shape & Sub-Sampling & Activation\\
\noalign{\smallskip}\hline\noalign{\smallskip}
\texttt{conv1} & $2\times216\times128$ & $2\times16\times5\times5$ & $2\times1$ & \texttt{ReLU}\\
\texttt{max-pool1} & $16\times108\times128$ &  & $2\times2$ & \\
\texttt{conv2} & $16\times54\times64$ &  $16\times32\times3\times3$ & & \texttt{ReLU}\\
\texttt{max-pool2} & $32\times54\times64$ &  & $2\times2$ & \\
\texttt{conv3} & $32\times27\times32$ & $32\times64\times3\times3$ & & \texttt{ReLU} \\
\texttt{max-pool3} & $64\times27\times32$ &  & $2\times2$ & \\
\texttt{conv4} & $64\times13\times16$ & $64\times64\times3\times3$ & & \texttt{ReLU}\\
\texttt{max-pool4} & $64\times13\times16$ &  & $2\times2$ &\\
\texttt{conv5} & $64\times6\times8$ & $64\times128\times3\times3$ & & \texttt{ReLU}\\
\texttt{max-pool5} & $128\times6\times8$ &  & $2\times2$ & \\
\texttt{conv61} & $128\times3\times4$  & $128\times256\times3\times3$ & & \texttt{ReLU}\\
\texttt{conv62} & $256\times3\times4$  & $256\times256\times1\times1$ & & \texttt{ReLU} \\
\texttt{gap} & $256$ &  &  \\
\texttt{fc-feature} & $256$ & $256\times256$ & & \texttt{ReLU} \\
\texttt{dropout} & $256$ &  &  \\
\texttt{fc-output} & $256$ & learning source specific & & \texttt{Softmax} \\
\noalign{\smallskip}\hline
\end{tabular}
\end{table} 

As the deep base architecture for feature representation learning, we choose a Convolutional Neural Network (CNN) architecture inspired by~\cite{Simonyan2014VeryRecognition}, as described in Fig.~\ref{fig:base_arch} and Table~\ref{tab:netarch}.

The CNN is one of the most popular architectures in many music-related machine learning tasks~\cite{Oord2013DeepRecommendation,Choi2016AutomaticNetworks,Han2016DeepMusic,Schluter2016LearningExamples,DBLP:conf/icassp/HersheyCEGJMPPS17,DBLP:conf/nips/LeePLN09,Dieleman2011Audio-basedNetwork,DBLP:conf/icmla/HumphreyB12,DBLP:conf/interspeech/NakashikaGT12,DBLP:conf/ismir/UllrichSG14,DBLP:conf/mlsp/Piczak15,DBLP:conf/ica/SimpsonRP15,DBLP:conf/interspeech/PhanHMM16,DBLP:conf/cbmi/PonsLS16,DBLP:conf/fedcsis/StasiakM16,DBLP:conf/icassp/SuZZG16}. Many of these works adopt an architecture having cascading blocks of 2-dimensional filters and max-pooling, derived from well-known works in image recognition~\cite{Simonyan2014VeryRecognition,Krizhevsky2012ImageNetNetworks}. Although variants of CNN using 1-dimensional filters also were suggested by~\cite{Dieleman2014END-TO-ENDAUDIO,Oord2016WaveNet:Audio,Aytar2016SoundNet:Video,Jaitly2011LEARNINGHinton} to learn features directly from a raw audio signal in an end-to-end manner, not many works managed to use them on music classification tasks successfully~\cite{Lee2017Sample-LevelWaveforms}.\par

The main difference between the base architecture and~\cite{Simonyan2014VeryRecognition} is the use of Global Average Pooling (GAP) and the Batch Normalization (BN) layers. BN is applied to accelerate the training and stabilize the internal covariate shift for every convolution layer and the \texttt{fc-feature} layer~\cite{Ioffe}. Also, global spatial pooling is adopted as the last pooling layer of the cascading convolution blocks, which is known to effectively summarize the spatial dimensions both in the image~\cite{he2016deep} and music domain~\cite{Han2016DeepMusic}. We also applied the approach to ensure the \texttt{fc-feature} layer not to have a huge number of parameters.

We applied the Rectified Linear Unit (ReLU)~\cite{Nair2010RectifiedMachines} to all convolution layers and the \texttt{fc-feature} layer. For the \texttt{fc-output} layer, softmax activation is used. For each convolution layer, we applied zero-padding such that the input and the output have the same spatial shape. As for the regularization, we choose to apply drop-out~\cite{Srivastava2014Dropout:Overfitting} on the \texttt{fc-feature} layer. We added $L2$ regularization across all the parameters with the same weight $\lambda=10^{-6}$.

\subsubsection{Audio Preprocessing}
\label{dl_specifications:audiopreproc}
We aim to learn a music representation from as-raw-as-possible input data to fully leverage the capability of the neural network. For this purpose, we use the dB-scale mel-scale magnitude spectrum of an input audio fragment, extracted by applying 128-band mel-filter banks on the Short-Time Fourier Transform (STFT). mel-spectrograms have generally been a popular input representation choice for CNNs applied in music-related tasks~\cite{Nam2012LearningRetrieval,Hamel2013TRANSFERSIMILARITY,Oord2013DeepRecommendation,Choi2016AutomaticNetworks,Choi2017TransferTasks,Han2016DeepMusic}; besides, it also was reported recently that their frequency-domain summarization, based on psycho-acoustics, is efficient and not easily learnable through data-driven approaches~\cite{Choi2017ATagging,Doerfler2017BasicDesign}. We choose a 1024-sample window size and 256-sample hop size, translating to about 46 ms and 11.6 ms respectively for a sampling rate of 22 kHz. We also applied standardization to each frequency band of the mel spectrum, making use of the mean and variance of all individual mel spectra in the training set.

\subsubsection{Sampling}
\label{dl_specifications:sampling}
During the learning process, in each iteration, a random batch of songs is selected. Audio corresponding to these songs originally is 30 seconds in length; for computational efficiency, we randomly crop 2.5 seconds out of each song each time. Keeping stereo channels of the audio, the size of a single input tensor $X^*$ we used for the experiment ended up with $2\times216\times128$, where the first dimension indicates number of channels, and following dimensions mean time steps and mel-bins, respectively. Along with the computational efficiency, a number of literatures in MIR field reported that using a small chunk of the input not only inflates the dataset, but also shows good performance on the high-level tasks such as music auto-tagging~\cite{Lee2017Sample-LevelWaveforms,Han2016DeepMusic,Dieleman2014END-TO-ENDAUDIO}. For the \textit{self} case, we generate batches with equal numbers of songs for both membership categories in $y_{self}$.\par

\begin{figure}[htp]
% 	\vspace*{-0.1in}
% 	\captionsetup{farskip=0pt}
    \centering
    \includegraphics[height=0.33\textheight]{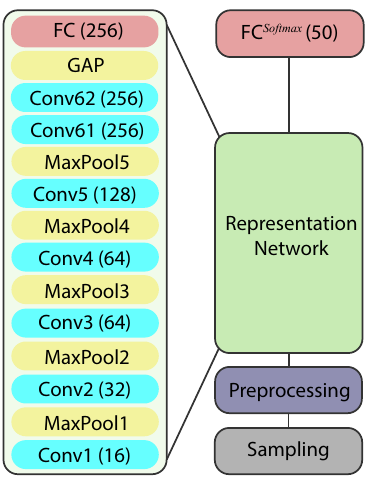}%
    \caption{Default CNN architecture for supervised single-source representation learning. Details of the Representation Network are presented at the left of the global architecture diagram. The numbers inside the parentheses indicate either the number of filters, or the number of units with respect to the type of layer.}
%     \vspace{-1em}
    \label{fig:base_arch}
\end{figure}

\subsection{Multi-Source Architectures with Various Degrees of Shared Information}
\label{dl_specifications:fusion}

\begin{figure}[htp]
    \centering
    \subfloat[SS-R: Base setup.]{%
        \includegraphics[height=0.33\textheight]{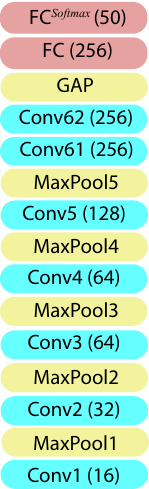}%
        \label{fig:base}%
        }%
    \hfill%
        \subfloat[MSS-CR: Concatenation of multiple independent SS-R networks.]{%
        \includegraphics[height=0.33\textheight]{graphics/split_base_black_no_margin.pdf}%
        \includegraphics[height=0.33\textheight]{graphics/split_base_black_no_margin.pdf}%
        \label{fig:mst_cr}%
        }%
    \hfill%
    \subfloat[MS-CR@2: network branches to source-specific layers from 2nd convolution layer.]{%
        \includegraphics[height=0.33\textheight]{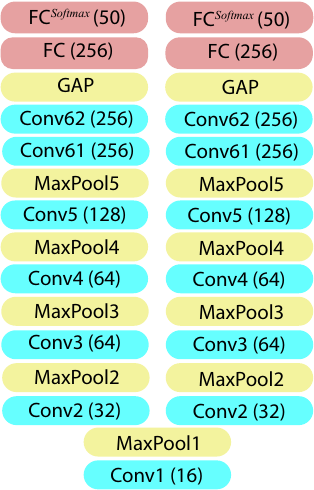}%
        \label{fig:split_2}%
        }%
    \hfill%
    \subfloat[MS-CR@4: network branches to source-specific layers from 4th convolution layer.]{%
        \includegraphics[height=0.33\textheight]{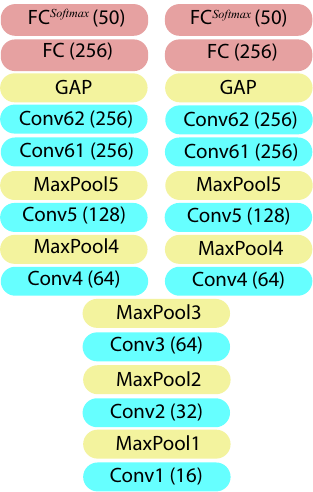}%
        \label{fig:split_4}%
        }%
    \hfill%
    \subfloat[MS-CR@6:  network branches to source-specific layers from 6th convolution layer.]{%
        \includegraphics[height=0.33\textheight]{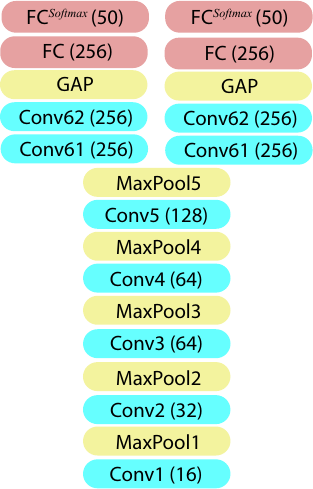}%
        \label{fig:split_6}%
        }%
    \hfill%
        \subfloat[MS-SR@FC: heavily shared network, source-specific branching only at final FC layer.]{%
        \includegraphics[height=0.33\textheight]{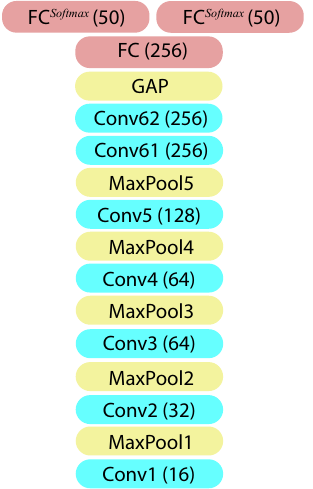}%
        \label{fig:split_fc}%
        }%
    \hfill
    \caption{The various model architectures considered in the current work. Beyond single-source architectures, multi-source architectures with various degrees of shared information are studied. For simplification, multi-source cases are illustrated here for two sources. The \texttt{fc-feature} layer from which representations will be extracted is the FC(256) layer in the illustrations (see Table~\ref{tab:netarch}).}
    \label{fig:split}
\end{figure}

When learning a music representation based on various available learning sources, different strategies can be taken regarding the choice of architecture. We will investigate the following setups:

\begin{itemize}
\item{
As a base case, a \emph{\textbf{Single-Source Representation (SS-R)}} can be learned for a single source only. As mentioned earlier, this would be the typical strategy leading to pre-trained networks, that later would be used in transfer learning. In our case, our base architecture from Section~\ref{dl_specifications:base_architecture} and Fig.\ \ref{fig:base_arch} will be used, for which the layers in the Representation Network also are illustrated in Fig.\ \ref{fig:base}. Out of the \texttt{fc-feature} layer, a $d$-dimensional representation is obtained.
}
\item{
If multiple perspectives on the same content, as reflected by the multiple learning labels, should also be reflected in the ultimate learned representation, one can learn \emph{SS-R} representations for each learning source, and simply concatenate them afterwards. With $d$ dimensions per source and $m$ sources, this leads to a $d \times m$ \emph{\textbf{Multiple Single-Source Concatenated Representation (MSS-CR)}}. In this case, independent networks are trained for each of the sources, and no shared knowledge will be transferred between sources. A layer setup of the corresponding Representation Network is illustrated in Fig.\ \ref{fig:mst_cr}.
}
\item{
When applying MTL learning strategies, the deep architecture should involve shared knowledge layers, before branching out to various individual learning sources, whose learned representations will be concatenated in the final $d \times m$-dimensional representation. We call these \emph{\textbf{Multi-Source Concatenated Representations (MS-CR)}}. As the branching point can be chosen at different stages, we will investigate the effect of various prototypical branching point choices: at the second convolution layer (\emph{MS-CR@2}, Fig.~\ref{fig:split_2}), the fourth convolution layer (\emph{MS-CR@4}, Fig.~\ref{fig:split_4}), and the sixth convolution layer (\emph{MS-CR@6}, Fig.~\ref{fig:split_6}).
The later the branching point occurs, the more shared knowledge the network will employ.
}
\item{
In the most extreme case, branching would only occur at the very last fully connected layer, and a \textbf{Multi-Source Shared Representation (MS-SR)} (or, more specifically, \emph{MS-SR@FC}) is learned, as illustrated in Fig.~\ref{fig:split_fc}. As the representation is obtained from the \texttt{fc-feature} layer, no concatenation takes place here, and a $d$-dimensional representation is obtained.
}
\end{itemize}

A summary of these different representation learning architectures is given in Table~\ref{tab:fusion}. Beyond the strategies we choose, further approaches can be thought of to connect representations learned for different learning sources in neural network architectures. For example, for different tasks, representations can be extracted from different intermediate hidden layers, benefiting from the hierarchical feature encoding capability of the deep network~\cite{Choi2017TransferTasks}. However, considering that learned representations are usually taken from a specific fixed layer of the shared architecture, we focus on the strategies as we outlined above.\par

\begin{table}
\centering
\caption{Properties of the various categories of representation learning architectures.}
\label{tab:fusion}
\begin{tabular}{ccccc}
\hline\noalign{\smallskip}
 & Multi Source & Shared Network & Concatenation & Dimensionality\\
\noalign{\smallskip}\hline\noalign{\smallskip}
\textbf{SS-R}   & No & No & No & $d$ \\
\textbf{MSS-CR} & Yes & No & Yes & $d\times{m}$ \\
\textbf{MS-CR}  & Yes & Partial & Yes & $d\times{m}$ \\
\textbf{MS-SR}  & Yes & Yes & No & $d$ \\
\noalign{\smallskip}\hline
\end{tabular}
\end{table}

\subsection{MTL Training Procedure}
\label{dl_specifications:train}

\begin{algorithm}[h]
\nl Initialize $\Theta$: \{$\theta^{t}$, $\theta^{s}$\} randomly\;
\nl \For{epoch in 1...N}{
\nl 	\For{iteration in 1...L}{
\nl			Pick a learning source $t$ randomly\;
\nl			Pick batch of samples from learning source $t$\;
	   		($X_l$, $X_r$) for \textit{self}\;
       		$X$ otherwise\;
\nl    		Derive learning label $z_{t}$\;
\nl	   		Sub-sample chunk $X^*$ from track $X$\;
\nl    		Forward-pass:\;
       		$\mathcal{L}(y_{self}, \Theta, X_l^*, X_r^*)=$Eq. \ref{eq:self} for \textit{self}\;
	   		$\mathcal{L}(z_{t}, \Theta, X^*)=$Eq. \ref{eq:2} otherwise\;
\nl    		Backward-pass: $\nabla(\Theta)$\;
\nl    		Update model: $\Theta \gets \Theta - \epsilon \nabla(\Theta)$\;
	}
  }
\caption{{Training a Multi-Source CNN} \label{Algorithm}}
\label{alg:train}
\end{algorithm}

Similar to~\cite{Weston2011Multi-TaskingRetrieval,Liu2015Multi-taskSelection}, we choose to train the MTL models with a stochastic update scheme as described in Algorithm~\ref{alg:train}. At every iteration, a learning source is selected randomly. After the learning source is chosen, a batch of observation-label pairs $(X, z_{t})$ is drawn. For the audio previews belonging to the songs within this batch, an input representation $X^*$ is cropped randomly from its super-sample $X$. The updates of the parameters $\Theta$ are conducted through back-propagation using the Adam algorithm~\cite{Kingma2014Adam:Optimization}. For each neural network we train, we set $L=lm$, where $l$ is the number of iterations needed to visit all the training samples with fixed batch size $b=128$, and $m$ is the number of learning sources used in the training. Across the training, we used a fixed learning rate $\epsilon=0.00025$. After a fixed number of epochs $N$ is reached, we stop the training.\par

\subsection{Implementation Details}
\label{dl_specifications:imple}

We used \textit{PyTorch}~\cite{paszke2017automatic} to implement the CNN models and parallel data serving. For evaluation of models and cross-validation, we made extensive use of functionality in \textit{Scikit-Learn}~\cite{Pedregosa2012Scikit-learn:Python}. Furthermore, \textit{Librosa}~\cite{Mcfee2015Librosa:Python} was used to process audio files and its raw features including mel spectrograms. The training is conducted with 8 Graphical Processing Unit (GPU) computation nodes, composed of 2 NVIDIA GRID K2 GPUs and 6 NVIDIA GTX 1080Ti GPUs.\par

\begin{figure}[htp]
\centering
\includegraphics[width=1\textwidth]{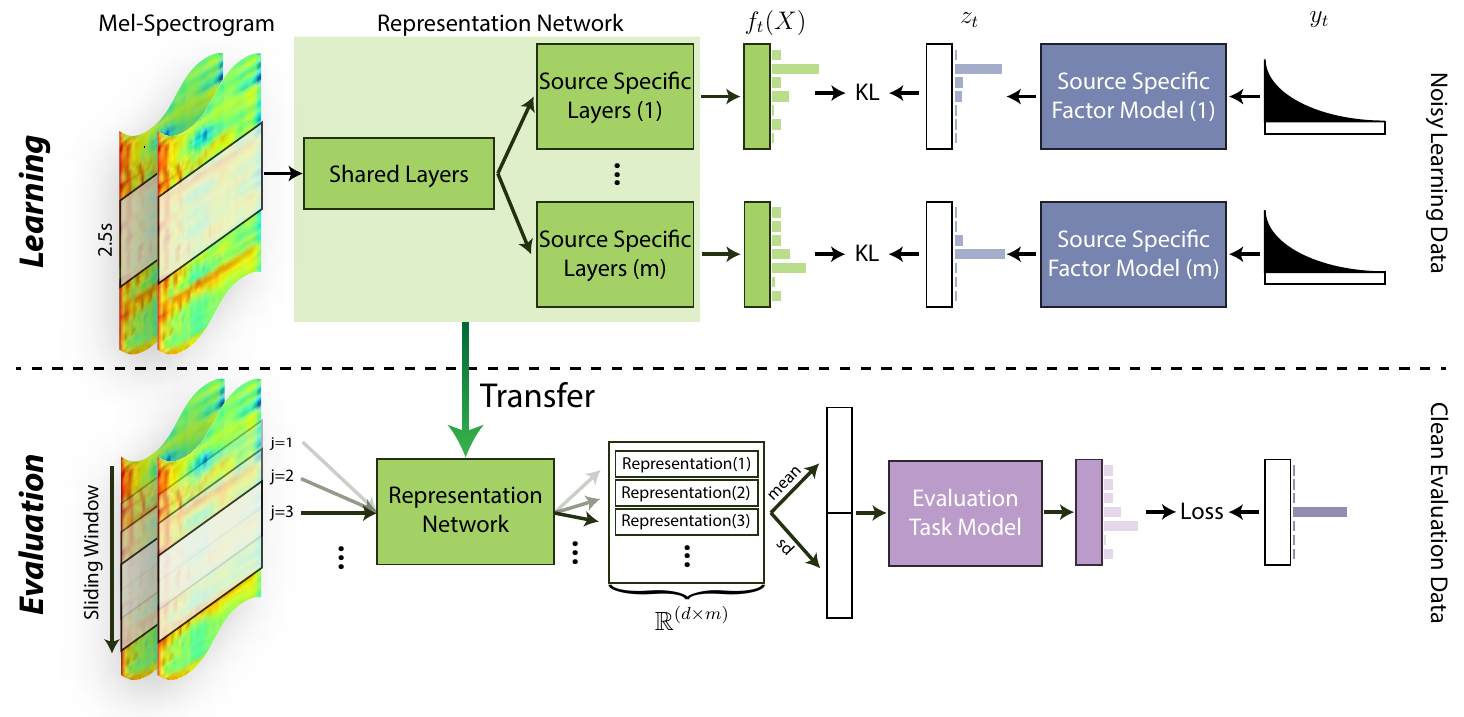}
\caption{Overall system framework. The first row of the figure illustrates the learning scheme, where the representation learning is happening by minimizing the KL divergence between the network inference $f_t(X)$ and the preprocessed learning label $z_t$. The preprocessing is conducted by the blue blocks which transform the original noisy labels $y_t$ to $z_t$, reducing noise and summarizing the high-dimensional label space into a smaller latent space. The second row describes the entire evaluation scenario. The representation is first extracted from the representation network, which is transferred from the upper row. The sequence of representation vectors is aggregated as the concatenation of their means and standard deviations. The purple block indicates a machine learning model employed to evaluate the representation's effectiveness.}
\label{fig:framework}
\end{figure}

\section{Evaluation}
\label{eval}

% \julian{todo}
So far, we discussed the details regarding the learning phase of this work, which corresponds to the upper row of Fig.~\ref{fig:framework}. This included various choices of sources for the representation learning, and various choices of architecture and fusion strategies. In this section, we present the evaluation methodology we followed, as illustrated in the second row of Fig.~\ref{fig:framework}. First, we will discuss the chosen target tasks and datasets in Section~\ref{eval:tasks}, followed in Section~\ref{eval:baseline} by the baselines against which our representations will be compared. Section~\ref{eval:expdesign} explains our experimental design, and finally we discuss the implementation of our evaluation experiments in Section~\ref{eval:imple}.

\subsection{Target Datasets}
\label{eval:tasks}

In order to gain insight into the effectiveness of learned representations with respect to multiple potential future tasks, we consider a range of \emph{target datasets}. In this work, our target datasets are chosen to reflect various semantic properties of music, purposefully chosen semantic biases, or popularity in the MIR literature. Furthermore, the representation network should not be configured or learned to explicitly solve the chosen target datasets.

While for the learning sources, we could provide categorizations on where and how the learning labels were derived, and also consider algorithmic outcomes as labels, existing popular research datasets mostly fall in the \textit{Professional} or \textit{Crowd} categories. In our work, we choose 7 evaluation datasets commonly used in MIR research, which reflect three conventional types of MIR tasks, namely classification, regression and recommendation:

\begin{table}[h]
\centering
\caption{Properties of target datasets used in our experiments. Because of time constraints, we sampled the Lastfm dataset as described in Section~\ref{eval:tasks}; the original size appears between parentheses. In case particular data splits are defined by an original author or follow up study, we apply the same split, including the reference in which the split is introduced. Otherwise, we applied either a random split stratified by the label (Ballroom), or simple filtering based on reported faulty entries (IRMAS).}
\label{tab:extertask}
\begin{tabular}{lrllll}
\hline\noalign{\smallskip}
Task & \multicolumn{2}{c}{Data} & \#Tracks & \#Class & Split Method \\
\noalign{\smallskip}\hline\noalign{\smallskip}
Classification & FMA\cite{Defferrard2016FMA:Analysis} & Genre & 25,000 & 16 & Artist Filtered~\cite{Defferrard2016FMA:Analysis} \\
Classification & GTZAN\cite{Tzanetakis2002MusicalIEEE} & Genre & 1,000 & 10 & Artist Filtered~\cite{DBLP:journals/tmm/KereliukSL15} \\
Classification & Ext. Ballroom\cite{Gouyon2006AnAlgorithms,DBLP:conf/mlsp/MarchandP16} & Genre & 3,390 & 13 & N/A \\
Classification & IRMAS\cite{Bosch2012ASignals} & Instrument & 6,705 & 11 & Song Filtered \\
Regression & Music Emotion\cite{Soleymani20131000Music} & Arousal & 744 &  & Genre Stratified\cite{Soleymani20131000Music}\\
Regression & Music Emotion\cite{Soleymani20131000Music} & Valence & 744 &  & Genre Stratified\cite{Soleymani20131000Music} \\
Recommendation & Lastfm\mbox{*}\cite{Celma:Springer2010} & Listening Count  & 27,093 (961,416) & & N/A \\
\noalign{\smallskip}\hline
\end{tabular}
\end{table}

\begin{itemize}
\item \textbf{\textit{Classification.}} Different types of classification tasks exist in MIR. In our experiments, we consider several datasets used for genre classification and instrument classification.

For genre classification, we chose the GTZAN~\cite{Tzanetakis2002MusicalIEEE} and FMA~\cite{Defferrard2016FMA:Analysis} datasets as main exemplars. Even though GTZAN is known for its caveats~\cite{Sturm2014TheRetrieval}, we deliberately used it, because its popularity can be beneficial when comparing with previous and future work. We note though that there may be some overlap between the tracks of GTZAN and the subset of the MSD we use in our experiments; the extent of this overlap is unknown, due to the lack of a confirmed and exhaustive track listing of the GTZAN dataset. We choose to use a fault-filtered data split for the training and evaluation, which is suggested in~\cite{DBLP:journals/tmm/KereliukSL15}. The split originally includes a training, validation and evaluation split; in our case, we also included the validation split as training data.

Among the various packages provided by the FMA, we chose the top-genre classification task of FMA-Medium~\cite{Defferrard2016FMA:Analysis}. This is a classification dataset with an unbalanced genre distribution. We used the data split provided by the dataset for our experiment, where the training is validation set are combined as the training.

Considering another type of genre classification, we selected the Extended Ballroom dataset~\cite{Gouyon2006AnAlgorithms, DBLP:conf/mlsp/MarchandP16}. Because the classes in this dataset are highly separable with regard to their BPM~\cite{Sturm2016TheSystems}, we specifically included this `purposefully biased' dataset as an example of how a learned representation may effectively capture temporal dynamics properties present in a target dataset, as long as learning sources also reflected these properties. Since no pre-defined split is provided or suggested by other literature, we used stratified random sampling based on the genre label.

The last dataset we considered for classification is the training set of the IRMAS dataset~\cite{Bosch2012ASignals}, which consists of short music clips annotated with the predominant instruments present in the clip. Compared to the genre classification task, instrument classification is generally considered as less subjective, requiring features to separate timbral characteristics of the music signal as opposed to high-level semantics like genre. We split the dataset to make sure that observations from the same music track are not split into training and test set.

As performance metric for all these classification tasks, we used classification accuracy.

\item \textbf{\textit{Regression.}} As exemplars of regression tasks, we evaluate our proposed deep representations on the dataset used in the MediaEval Music Emotion prediction task~\cite{Soleymani20131000Music}. It contains frame-level and song-level labels of a two-dimensional representation of emotion, with valence and arousal as dimensions~\cite{Posner2005ThePsychopathology}. Valence is related to the positivity or negativity of the emotion, and arousal is related to its intensity~\cite{Soleymani20131000Music}. The song-level annotation of the V-A coordinates was used as the learning label. In similar fashion to the approach taken in~\cite{Choi2017TransferTasks}, we trained separate models for the two emotional dimensions. As for the dataset split, we used the split provided by the dataset, which is done by the random split stratified by the genre distribution.

As evaluation metric, we measured the coefficient of determination $R^{2}$ of each model.

\item \textbf{\textit{Recommendation.}} Finally, we employed the `Last.fm - 1K users' dataset~\cite{Celma:Springer2010} to evaluate our representations in the context of a content-aware music recommendation task (which will be denoted as \emph{Lastfm} in the remaining of the paper). This dataset contains 19 million records of listening events across $961,416$ unique tracks collected from $992$ unique users. In our experiments, we mimicked a cold-start recommendation problem, in which items not seen before should be recommended to the right users. For efficiency, we filtered out users who listened to less than $5$ tracks and tracks known to less than $5$ users.

As for the audio content of each track, we obtained the mapping between the MusicBrainz Identifier (MBID) with the Spotify identifier (SpotifyID) using the \texttt{MusicBrainz API}\footnote{\url{https://musicbrainz.org/}}. After cross-matching, we collected 30 seconds previews of all track using the \texttt{Spotify API}\footnote{\url{https://developer.spotify.com/documentation/web-api/}}. We found that there is a substantial amount of missing mapping information between the SpotifyID and MBID in the \texttt{MusicBrainz} database, where only approximately 30\% of mappings are available. Also, because of the substantial amount of inactive users and unpopular tracks in the dataset, we ultimately acquired a dataset of $985$ unique users and $27,093$ unique tracks with audio content.

Similar to \cite{Liang2014Content-AwareNetworks}, we considered the \textit{outer matrix} performance for un-introduced songs; in other words, the model's recommendation accuracy on the items newly introduced to the system~\cite{Liang2014Content-AwareNetworks}. This was done by holding out certain tracks when learning user models, and then predicting user preference scores based on all tracks, including those that were held out, resulting in a ranked track list per user. As evaluation metric, we consider Normalized Discounted Cumulative Gain ($nDCG@500$), only treating held-out tracks that were indeed liked by a user as relevant items. Further details on how hold-out tracks were chosen are given in Section~\ref{eval:imple}.
\end{itemize}

A summary of all evaluation datasets, their origins and properties, can be found in Table~\ref{tab:extertask}.

\subsection{Baselines}
\label{eval:baseline}

We examined three baselines to compare with our proposed representations:
\begin{itemize}
\item\textbf{\textit{Mel-Frequency Cepstral Coefficients (MFCC).}} These are some of the most popular audio representations in MIR research. In this work, we extract and aggregate MFCC following the strategy in~\cite{Choi2017TransferTasks}. In particular, we extracted 20 coefficients and also used their first- and second-order derivatives. After obtaining the sequence of MFCCs and its derivatives, we performed aggregation by taking the average and standard deviation over the time dimension, resulting in a 120-dimensional vector representation.

\item\textbf{\textit{Random Network Feature (Rand).}} We extracted the representation at the \texttt{fc-feature} layer without any representation network training. With random initialization, this representation therefore gives a random baseline for a given CNN architecture. We refer to this baseline as \textit{Rand}.

\item\textbf{\textit{Latent Representation from Music Auto-Tagger (Choi).}} The work in~\cite{Choi2017TransferTasks} focused on a music auto-tagging task, and can be considered as yielding a state-of-the-art deep music representation for MIR. While the model's focus on learning a representation for music auto-tagging can be considered as our \textit{SS-R} case, there are a number of issues that complicate direct comparisons between this work and ours. First, the network in~\cite{Choi2017TransferTasks} is trained with about 4 times more data samples than in our experiments. Second, it employed a much smaller network than our architecture. Further, intermediate representations were extracted, which is out of the scope of our work, as we only consider representations at the \texttt{fc-feature} layer. Nevertheless, despite these caveats, the work still is very much in line with ours, making it a clear candidate for comparison. Throughout the evaluation, we could not fully reproduce the performance reported in the original paper~\cite{Choi2017TransferTasks}. When reporting our results, we therefore will report the performance we obtained with the published model, referring to this as \textit{Choi}.
% When reporting our results, we therefore will report both the performance we obtained with the published model, referring to this as \textit{Choi}, and the one reported in the original paper, referred to as \textit{Choi paper}.\julian{review this about the two choi versions}
\end{itemize}

\subsection{Experimental Design}
\label{eval:expdesign}

\begin{figure}
\centering
\includegraphics[scale=.4]{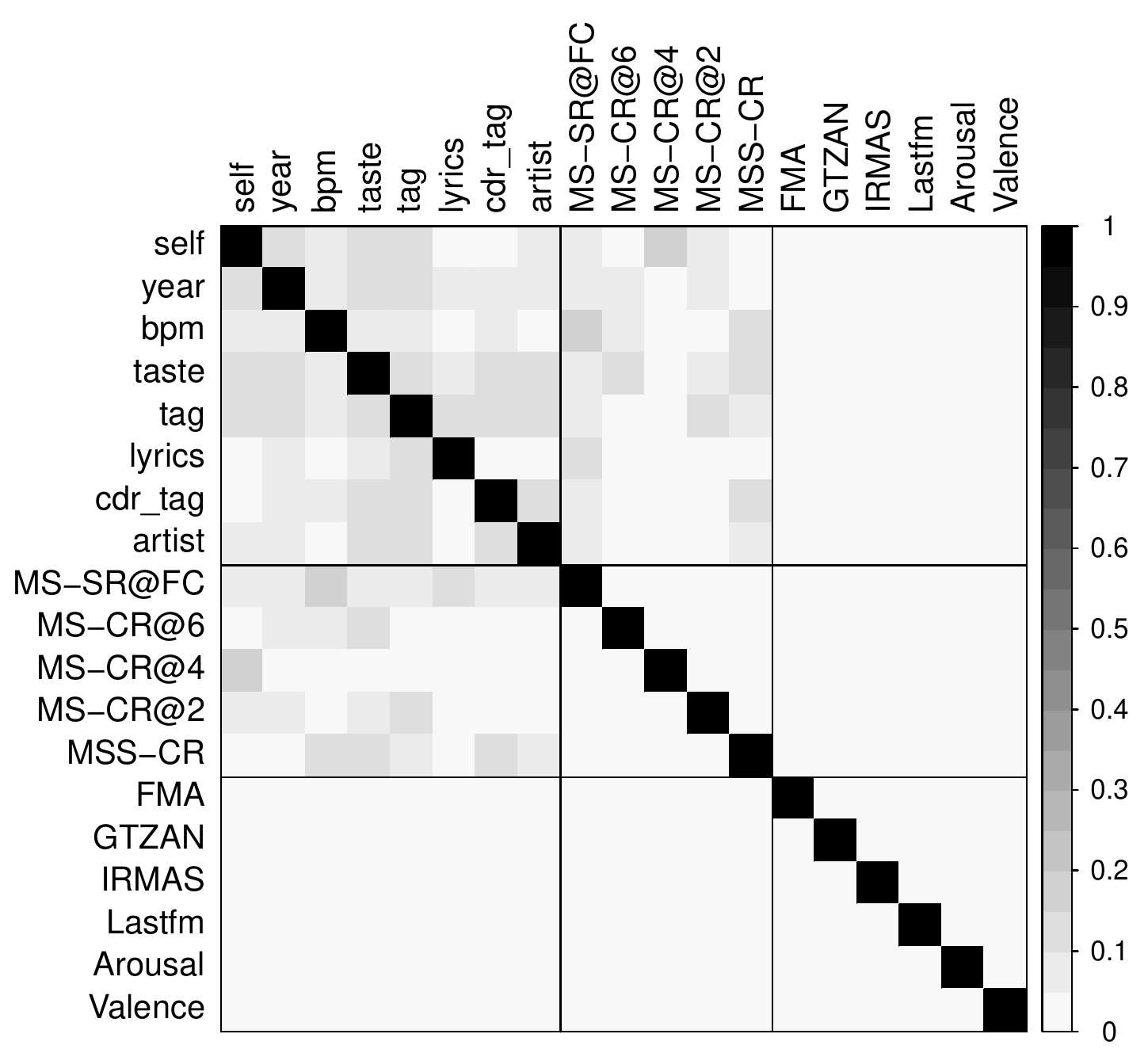}
\caption{Aliasing among main effects in the final experimental design.}
\label{fig:exp_design}
\end{figure}

In order to investigate our research questions, we carried out an experiment to study the effect of the number and type of learning sources on the effectiveness of deep representations, as well as the effect of the various architectural learning strategies described in Section~\ref{dl_specifications:fusion}. For the experimental design we consider the following factors:
\begin{itemize}
  \item Representation strategy, with 6 levels: \emph{SS-R}, \emph{MS-SR@FC}, \emph{MS-CR@6}, \emph{MS-CR@4}, \emph{MS-CR@2}, and \emph{MSS-CR}).
  \item 8 2-level factors indicating the presence or not of each of the 8 learning sources: \emph{self}, \emph{year}, \emph{bpm}, \emph{taste}, \emph{tag}, \emph{lyrics}, \emph{cdr\_tag} and \emph{artist}.
  \item Number of learning sources present in the learning process (1 to 8). Note that this is actually calculated as the sum of the eight factors above.
  \item Target dataset, with 7 levels: Ballroom, FMA, GTZAN, IRMAS, Lastfm, Arousal and Valence.
\end{itemize}
Given a learned representation, fitting dataset-specific models is much more efficient than learning the representation, so we decided to evaluate each representation on all 7 target datasets. The experimental design is thus restricted to combinations of representation and learning sources, and for each such combination we will produce 7 observations. However, given the constraint of \emph{SS-R} relying on a single learning source, that there is only one possible combination for n = 8 sources, as well as the high unbalance in the number of sources\footnote{For instance, from the 255 possible combinations of up to 8 sources, there are 70 combinations of $n=4$ sources, but 28 with $n=2$, or only 8 for $n=7$. Simple random sampling from the 255 possible combinations would lead to a very unbalanced design, that is, a highly non-uniform distribution of observation counts across the levels of the factor ($n$ in this case). A balanced design is desired to prevent aliasing and maximize statistical power. See section 15.2 in~\cite{Montgomery2012design} for details on unbalanced designs.}, we proceeded in three phases:
\begin{enumerate}
\item We first trained the \emph{SS-R} representations for each of the 8 sources, and repeated 6 times each. This resulted in 48 experimental runs.
\item We then proceeded to train all five multi-source strategies with all sources, that is, $n=8$. We repeated this 5 times, leading to 25 additional experimental runs.
\item Finally, we ran all five multi-source strategies with $n=2,\dots,7$. The full design matrix would contain 5 representations and 8 sources, for a total of 1,230 possible runs. Such an experiment was unfortunately infeasible to run exhaustively given available resources, so we decided to follow a fractional design. However, rather than using a pre-specified optimal design with a fixed amount of runs~\cite{Goos2011optimal}, we decided to run sequentially for as long as time would permit us, generating at each step a new experimental run on demand in a way that would maximize desired properties of the design up to that point, such as balance and orthogonality\footnote{An experimental design is orthogonal if the effects of any factor balance out across the effects of the other factors. In a non-orthogonal design effects may be aliased, meaning that the estimate of one effect is partially biased with the effect of another, the extent of which ranges from 0 (no aliasing) to 1 (full aliasing). Aliasing is sometimes referred to as confounding. See sections 8.5 and 9.5 in~\cite{Montgomery2012design} for details on aliasing.}.

We did this with the greedy Algorithm~\ref{alg:design}. From the set of still remaining runs $\mathcal{A}$, a subset $\mathcal{O}$ is selected such that the expected unbalance in the augmented design $\mathcal{B}\cup\{o\}$ is minimal. In this case, the unbalance of a design is defined as the maximum unbalance found between the levels of any factor, except for those already exhausted\footnote{For instance, let a design have 20 runs for \emph{SS-R}, 16 for \emph{MS-SR@FC}, and 18 for all other representations. The unbalance in the representation factor is thus $20-16=4$. The total unbalance of the design is defined as the maximum unbalance found across all factors.}. 
From $\mathcal{O}$, a second subset $\mathcal{P}$ is selected such that the expected aliasing in the augmented design is minimal, here defined as the maximum absolute aliasing between main effects\footnote{See section 2.3.7 in~\cite{Goos2011optimal} for details on how to compute an alias matrix.}. Finally, a run $p$ is selected at random from $\mathcal{P}$, the corresponding representation is learned, and the algorithm iterates again after updating $\mathcal{A}$ and $\mathcal{B}$. 

Following this on demand methodology, we managed to run another 352 experimental runs from all the 1,230 possible.
\end{enumerate}

\begin{algorithm}[h]
\nl Initialize $\mathcal{A}$ with all possible 1,230 runs to execute\;
\nl Initialize $\mathcal{B}\gets\emptyset$ for the set of already executed runs\;
\nl \While{time allows}{
\nl			Select $\mathcal{O}\subseteq \mathcal{A}$ s.t. $\forall o\in \mathcal{O}$, the unbalance in $\mathcal{B}\cup \{o\}$ is minimal\;
\nl     Select $\mathcal{P}\subseteq \mathcal{O}$ s.t. $\forall p\in \mathcal{P}$, the aliasing in $\mathcal{B}\cup \{p\}$ is minimal\;
\nl			Select $p\in \mathcal{P}$ at random\;
\nl     Update $\mathcal{A}\gets \mathcal{A}-\{p\}$\;
\nl     Update $\mathcal{B}\gets \mathcal{B}\cup\{p\}$\;
\nl     Learn the representation coded by $p$\;
}
\caption{Sequential generation of experimental runs.}
\label{alg:design}
\end{algorithm}

After going through the three phases above, the final experiment contained $48+25+352=425$ experimental runs, each producing a different deep music representation. We further evaluated each representation on all 7 target datasets, leading to a grand total of $42\times 7=2,97$5 datapoints. Fig.~\ref{fig:exp_design} plots the alias matrix of the final experimental design, showing that the aliasing among main factors is indeed minimal. The final experimental design matrix can be downloaded along with the rest of the supplemental material.

Each considered representation network was trained using the CNN representation network model from Section~\ref{dl_specifications}, based on the specific combination of learning sources and deep architecture as indicated by the experimental run. In order to reduce variance, we fixed the number of training epochs to $N = 200$ across all runs, and applied the same base architecture, except for the branching point. This entire training procedure took approximately 5 weeks with given computational hardware resources introduced in Section~\ref{dl_specifications:imple}.

\subsection{Implementation Details}
\label{eval:imple}

In order to assess how our learned deep music representations perform on the various target datasets, transfer learning will now be applied, to consider our representations in the context of these new target datasets.

As a consequence, new machine learning pipelines are set up, focused on each of the target datasets. In all cases, we applied the pre-defined split if it is feasible. Otherwise, we randomly split the dataset in a 80\% training and 20\% test set. For every dataset, we repeated the training and evaluation for 5 times, using different train/test splits. In most of our evaluation cases, validation will take place on the test set; in case of the the recommendation problem, the test set represents a set of tracks to be held out during user model training, and re-inserted for validation. In all cases, we will extract representations from evaluation dataset audio as detailed in Section~\ref{eval:imple:feat_preproc}, and then learn relatively simple models based on them, as detailed in Section~\ref{eval:imple:model}. Employing the metrics as mentioned in the previous section, we will then take average performance scores over the 5 different train-test splits for final performance reporting.

\subsubsection{Feature Extraction and Preprocessing}
\label{eval:imple:feat_preproc}

Taking raw audio from the evaluation datasets as input, we take non-overlapping slices out of this audio with a fixed length of 2.5 seconds. Based on this, we apply the same preprocessing transformations as discussed in Section~\ref{dl_specifications:audiopreproc}. Then, we extract a deep representation from this preprocessed audio, employing the architecture as specified by the given experimental run. As in the case of Section~\ref{dl_specifications:fusion}, representations are extracted from the \texttt{fc-feature} layer of each trained CNN model. Depending on the choice of architecture, the final representation may consist of concatenations of representations obtained by separate representation networks.

Input audio may originally be (much) longer than 2.5 seconds; therefore, we aggregate information in feature vectors over multiple time slices by taking their \textit{mean} and \textit{standard deviation} values. As a result, we get a representation with averages per learned feature dimension, and another representation with standard deviations per feature dimension. These will be concatenated, as illustrated in Fig.~\ref{fig:framework}.\par

\subsubsection{Target Dataset-Specific Models}
\label{eval:imple:model}

As our goal is not to over-optimize dataset-specific performance, but rather perform a comparative analysis between different representations (resulting from different learning strategies), we keep the model simple, and use fixed hyper-parameter values for each model across the entire experiment. 

To evaluate the trained representations, we used different models according to the target dataset. For classification and regression tasks, we used Multi Layer Perceptron (MLP) model~\cite{DBLP:journals/ai/Hinton89}. More specifically, the MLP model has two hidden layers, whose dimensionality is $256$. As for the non-linearity, we choose ReLU~\cite{Nair2010RectifiedMachines} for all nodes, and the model is trained with ADAM optimization technique~\cite{Kingma2014Adam:Optimization} for 200 iterations. In evaluation, we used the \textit{Scikit-Learn}'s implementation for ease of distributed computing on multiple CPU computation nodes.

For the recommendation task, we choose a similar model as suggested in~\cite{Liang2014Content-AwareNetworks,Hu2008CollaborativeYifan}, in which the learning objective function $\mathcal{L}$ is defined as
\begin{equation} \label{eq:recsys}
\hat{U}, \hat{V}, \hat{W} = \argmin \; ||P-UV^{T}||_{C} + \frac{\lambda^{V}}{2}||V-XW|| + \frac{\lambda^{U}}{2}||U|| + \frac{\lambda^{W}}{2}||W||
\end{equation}
\noindent where $P\in\mathbb{R}^{u\times{i}}$ is a binary matrix indicating whether there is interaction between users $u$ and items $i$, $U\in\mathbb{R}^{u\times{r}}$ and $V\in\mathbb{R}^{i\times{r}}$ are $r$ dimensional user factors and item factors for the low-rank approximation of $P$. $P$ is derived from the original interaction matrix $R\in\mathbb{R}^{u\times{i}}$, which contains the number of interaction from users $u$ to items $i$, as follows:\par

\begin{equation}
P_{u, i} = 
\begin{cases}
    1, & \text{if } R_{u, i} > 0\\
    0 & \text{otherwise}
\end{cases}
\end{equation}

$W\in\mathbb{R}^{d\times{r}}$ is a free parameter for the projection from $d$-dimensional feature space to the factor space. $X\in\mathbb{R}^{i\times{d}}$ is the feature matrix where each row corresponds to a track. Finally, $||\cdot||_{C}$ is the Frobenious norm weighted by the confidence matrix $C\in\mathbb{R}^{u\times{i}}$, which controls the credibility of the model on the given interaction data, given as follows:\par

\begin{equation} \label{eq:recsys:confidence}
C = 1 + \alpha R
\end{equation}

where $\alpha$ controls credibility. As for hyper-parameters, we set $\alpha=0.1$, $\lambda^{V}=0.00001$, $\lambda^{U}=0.00001$, and $\lambda^{W}=0.1$, respectively. For the number of factors we choose $r=50$ to focus only on the relative impact of the representation over the different conditions. We implemented an update rule with the Alternating Least Squares (ALS) algorithm similar to~\cite{Liang2014Content-AwareNetworks}, and updated parameters during 15 iterations.

\section{Results and Discussion}
\label{res:intro}

In this section, we present results and discussion related to the proposed deep music representations. In Section \ref{res:single_multi_rep}, we will first compare the performance across the \emph{SS-R}s, to show how different individual learning sources work for each target dataset. Then, we will present general experimental results related to the performance of the multi-source representations. In Section \ref{res:task_num}, we discuss the effect of the number of learning sources exploited in the representation learning, in terms of their general performance, reliability, and model compactness. In Section \ref{res:single_vs_multi}, we discuss effectiveness of different representations in MIR. Finally, we present some initial evidence for multifaceted semantic explainability of the proposed MTDTL in Section~\ref{res:mulexpfac}.\footnote{For the reproducibility, we release all relevant materials including code, models and extracted features at \url{https://github.com/eldrin/MTLMusicRepresentation-PyTorch}.}

\subsection{Single-Source and Multi-Source Representation}
\label{res:single_multi_rep}

\begin{figure}
\centering
\includegraphics[scale=\figscale]{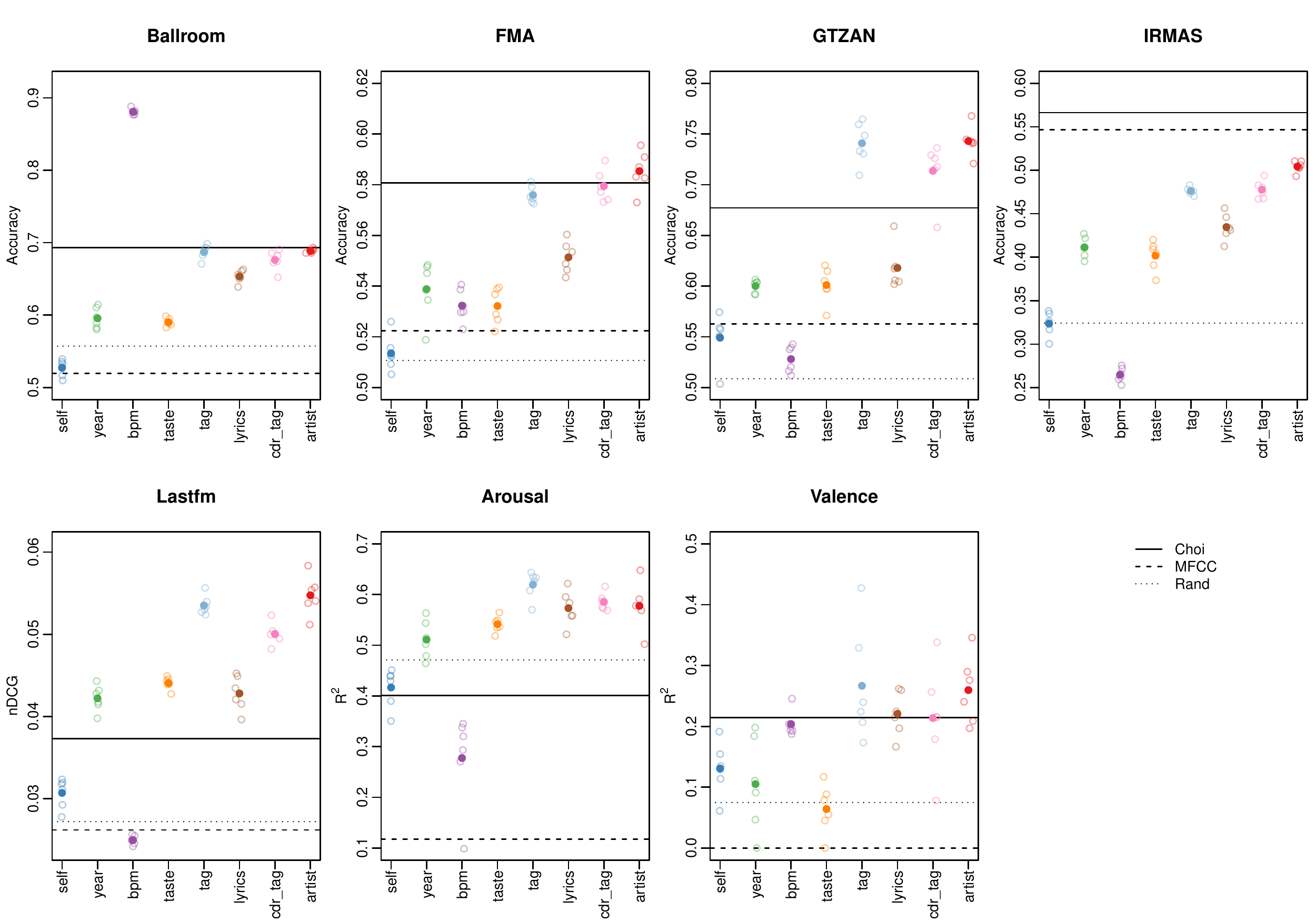}
\caption{Performance of single source representations. Each point indicates the performance of a representation learned from the single source. Solid points indicate the average performance per source. The baselines are illustrated as horizontal lines.}
\label{fig:single_task_rep}
\end{figure}

Fig.~\ref{fig:single_task_rep} presents the performance of \emph{SS-R} representations on each of the 7 target datasets. We can see that all sources tend to outperform the \textit{Rand} baseline on all datasets, except for a handful cases involving sources \emph{self} and \emph{bpm}. Looking at the top performing sources, we find that \emph{tag}, \emph{cdr\_tag} and \emph{artist} perform better or on-par with the most sophisticated baseline, \textit{Choi}, except for the IRMAS dataset. The other sources are found somewhere between these two baselines, except for datasets Lastfm and Arousal, where they perform better than \textit{Choi} as well. Finally, the \textit{MFCC} is generally outperformed in all cases, with the notable exception of the IRMAS dataset, where only \textit{Choi} performs better.

Zooming in to dataset-specific observed trends, the \emph{bpm} learning source shows a highly skewed performance across target datasets: it clearly outperforms all other learning sources in the Ballroom dataset, but it achieves the worst or second worst performance in the other datasets. As shown in~\cite{Sturm2016TheSystems}, this confirms that the Ballroom dataset is well-separable based on BPM information alone. Indeed, representations trained on the \emph{bpm} learning source seem to contain a latent representation close to the BPM of an input music signal. In contrast, we can see that the \emph{bpm} representation achieves the worst results in the Arousal dataset, where both temporal dynamics and BPM are considered as important factors determining the intensity of emotion.

On the IRMAS dataset, we see that all the \emph{SS-R}s perform worse than the \textit{MFCC} and \textit{Choi} baselines. Given that they both take into account low-level features, either by design or by exploiting low-level layers of the neural network, this suggests that predominant instrument sounds are harder to distinguish based solely on semantic features, which is the case of the representations studied here.

Also, we find that there is small variability for each \emph{SS-R} run within the training setup we applied. Specifically, in 50\% of cases we have within-\emph{SS-R} variability less than 15\% of the within-dataset variability. 90\% of the cases are within 30\% of the within-dataset variability.

\begin{figure}
\centering
\includegraphics[scale=\figscale]{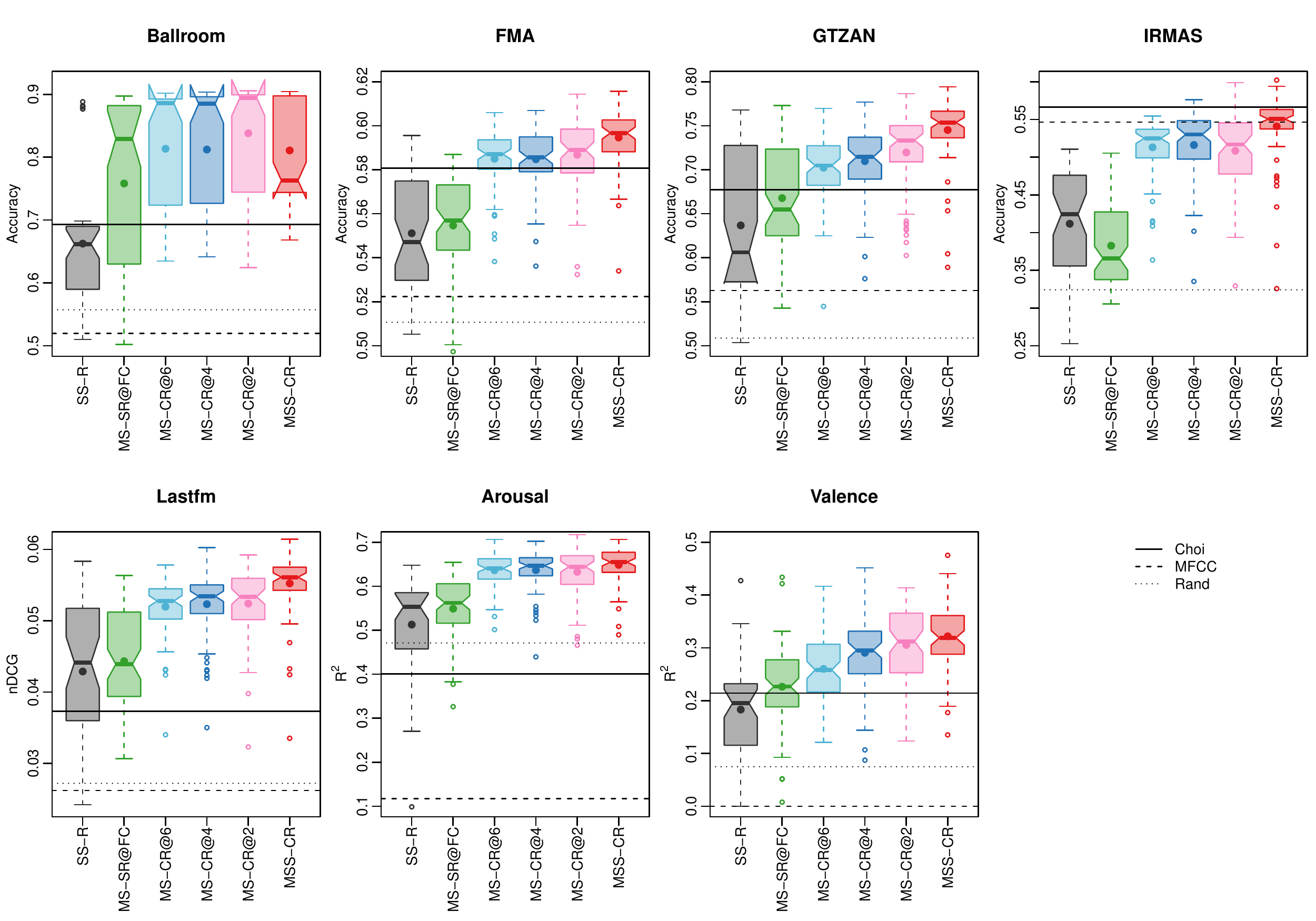}
\caption{Performance by representation strategy. Solid points represent the mean per representation. The baselines are illustrated as horizontal lines.}
\label{fig:overallperformance}
\end{figure}

We now consider how the various representations based on multiple learning sources perform, in comparison to those based on single learning sources. The boxplots in Fig.~\ref{fig:overallperformance} show the distributions of performance scores for each architectural strategy and per target dataset. For comparison, the gray boxes summarize the distributions depicted in Fig.~\ref{fig:single_task_rep}, based on the \emph{SS-R} strategy.
In general, we can see that these \emph{SS-R} obtain the lowest scores, followed by \emph{MS-SR@FC}, except for the IRMAS dataset. Given that these representations have the same dimensionality, these results suggest that adding a single source-specific layer on top of a heavily shared model may help improving the adaptability of the neural network models, especially when there is no prior knowledge regarding the well-matching learning sources for the target datasets.
The \emph{MS-CR} and \emph{MSS-CR} representations obtain the best results in general, which is somewhat expected because of their larger dimensionality.

\subsection{Effect of Number of Learning Sources and Fusion Strategy}
\label{res:task_num}

\begin{figure}
  \centering\includegraphics[scale=\figscale]{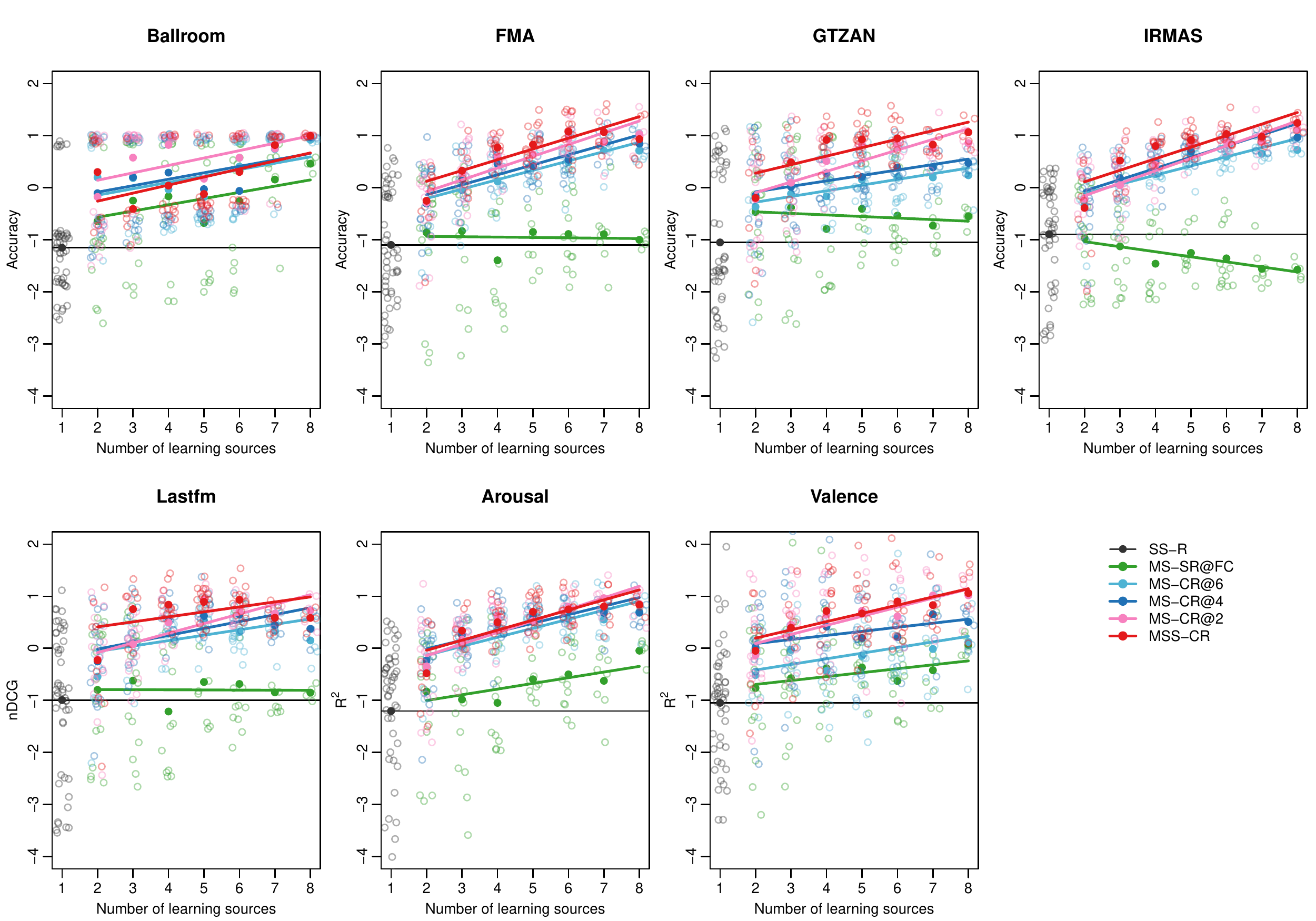}
  \caption{(Standardized) Performance by number of learning sources. Solid points represent the mean per architecture and number of sources. The black horizontal line marks the mean performance of the \emph{SS-R} representations. The colored lines show linear fits.}
  \label{fig:performance_by_n}
\end{figure}

While the plots in Fig.~\ref{fig:overallperformance} suggest that \emph{MSS-CR} and \emph{MS-CR} are the best strategies, the high observed variability makes this statement still rather unclear. In order to gain better insight of the effects of dataset, architecture strategies and number and type of learning sources, we further analyzed the results using a hierarchical or multilevel linear model on all observed scores~\cite{Gelman2006hierarchical}. The advantage of such a model is essentially that it accounts for the structure in our experiment, where observations nested within datasets are not independent.

By Fig.~\ref{fig:overallperformance} we can anticipate a very large dataset effect because of the inherently different levels of difficulty, as well as a high level of heteroskedasticity. We therefore analyzed standardized performance scores rather than raw scores. In particular, the $i$-th performance score $y_i$ is standardized with the within-dataset mean and standard deviation scores, that is, $y^*_i=(y_i - \bar{y}_{d[i]})/s_{d[i]}$, where $d[i]$ denotes the dataset of the $i$-th observation. This way, the dataset effect is effectively $0$ and the variance is homogeneous. In addition, this will allow us to compare the relative differences across strategies and number of sources using the same scale in all datasets.

We also transformed the variable $n$ that refers to the number of sources to $n^*$, which is set to $n^*=0$ for \emph{SS-R}s and to $n^*=n-2$ for the other strategies. This way, the intercepts of the linear model will represent the average performance of each representation strategy in its simplest case, that is, \emph{SS-R} ($n=1$) or non-\emph{SS-R} with $n=2$.
We fitted a first analysis model as follows:
\begin{align}
y^*_i &= \beta_{0r[i]d[i]} + \beta_{1r[i]d[i]}\cdot n^*_i + e_i &e_i&\sim N(0,\sigma^2_e) \label{eq:m11}\\
\beta_{0rd} &= \beta_{0r} + u_{0rd} &u_{0rd}&\sim N(0,\sigma^2_{0r}) \label{eq:m12}\\
\beta_{1rd} &= \beta_{1r} + u_{1rd} &u_{1rd}&\sim N(0,\sigma^2_{1r}) \label{eq:m13},
\end{align}
where $\beta_{0r[i]d[i]}$ is the intercept of the corresponding \underline{r}epresentation strategy within the corresponding \underline{d}ataset. Each of these coefficients is defined as the sum of a global fixed effect $\beta_{0r}$ of the representation, and a random effect $u_{0rd}$ which allows for random within-dataset variation\footnote{We note that hierarchical models do not fit each of the individual $u_{0rd}$ coefficients (a total of 42 in this model), but the amount of variability they produce, that is, $\sigma^2_{0r}$ (6 in total).}. This way, we separate the effects of interest (ie. each $\beta_{0r}$) from the dataset-specific variations (ie. each $u_{0rd}$). The effect of the number of sources is similarly defined as the sum of a fixed representation-specific coefficient $\beta_{1r}$ and a random dataset-specific coefficient $u_{1rd}$. Because the slope depends on the representation, we are thus implicitly modeling the interaction between strategy and number of sources, which can be appreciated in Fig.~\ref{fig:performance_by_n}, specially with \emph{MS-SR@FC}.

\begin{figure}
  \centering\includegraphics[scale=.4]{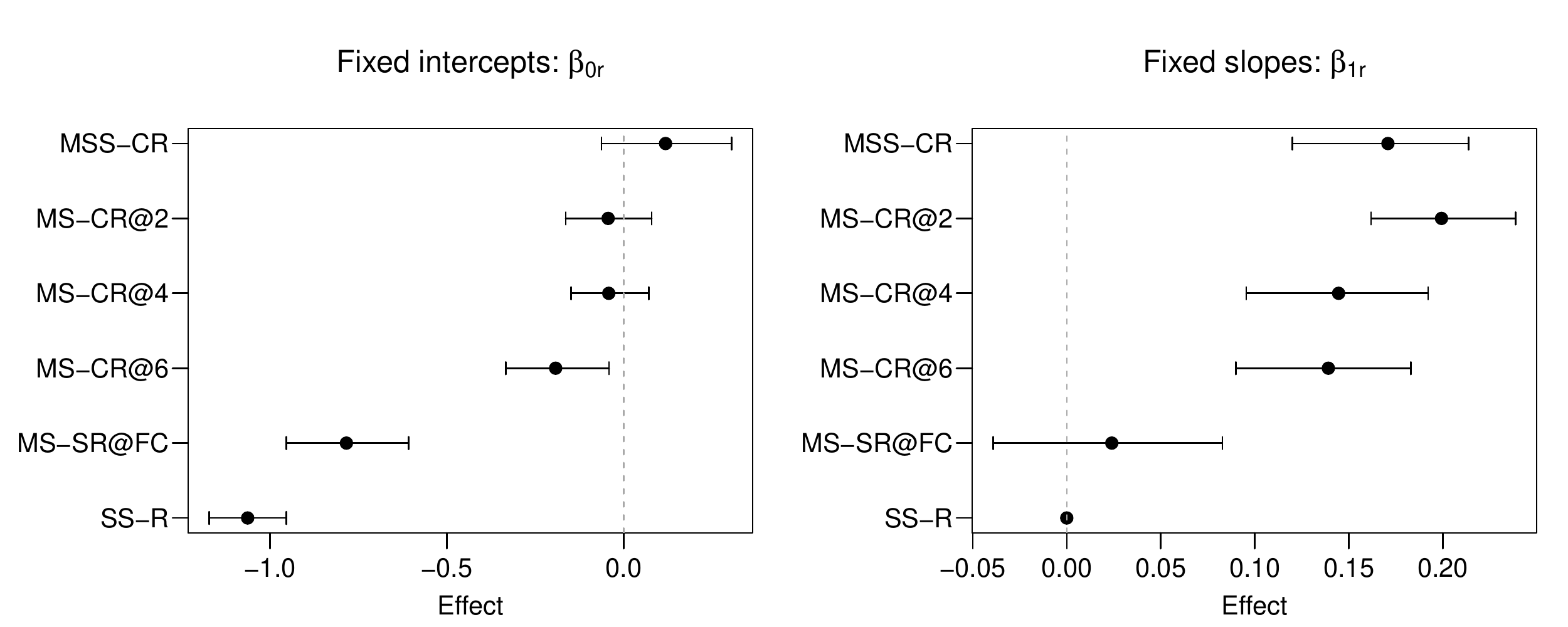}
  \caption{Fixed effects and bootstrap 95\% confidence intervals estimated for the first analysis model. The left plot depicts the effects of the representation strategy ($\beta_{0r}$ intercepts) and the right plot shows the effects of the number of sources ($\beta_{1r}$ slopes).}
  \label{fig:effects1}
\end{figure}

Fig.~\ref{fig:effects1} shows the estimated effects and bootstrap 95\% confidence intervals. The left plot confirms the observations in Fig.~\ref{fig:overallperformance}. In particular, they confirm that \emph{SS-R} performs significantly worse than \emph{MS-SR@FC}, which is similarly statistically worse than the others. When carrying out pairwise comparisons, \emph{MSS-CR} outperforms all other strategies except \emph{MS-CR@2} ($p=0.32$), which ourperforms all others except \emph{MS-CR@6} ($p=0.09$). The right plot confirms the qualitative observation from Fig.~\ref{fig:performance_by_n} by showing a significantly positive effect of the number of sources except for \emph{MS-SR@FC}, where it is not statistically different from 0. The intervals suggest a very similar effect in the best representations, with average increments of about $0.16$ per additional source ---recall that scores are standardized. 

\begin{figure}
  \centering\includegraphics[scale=.4]{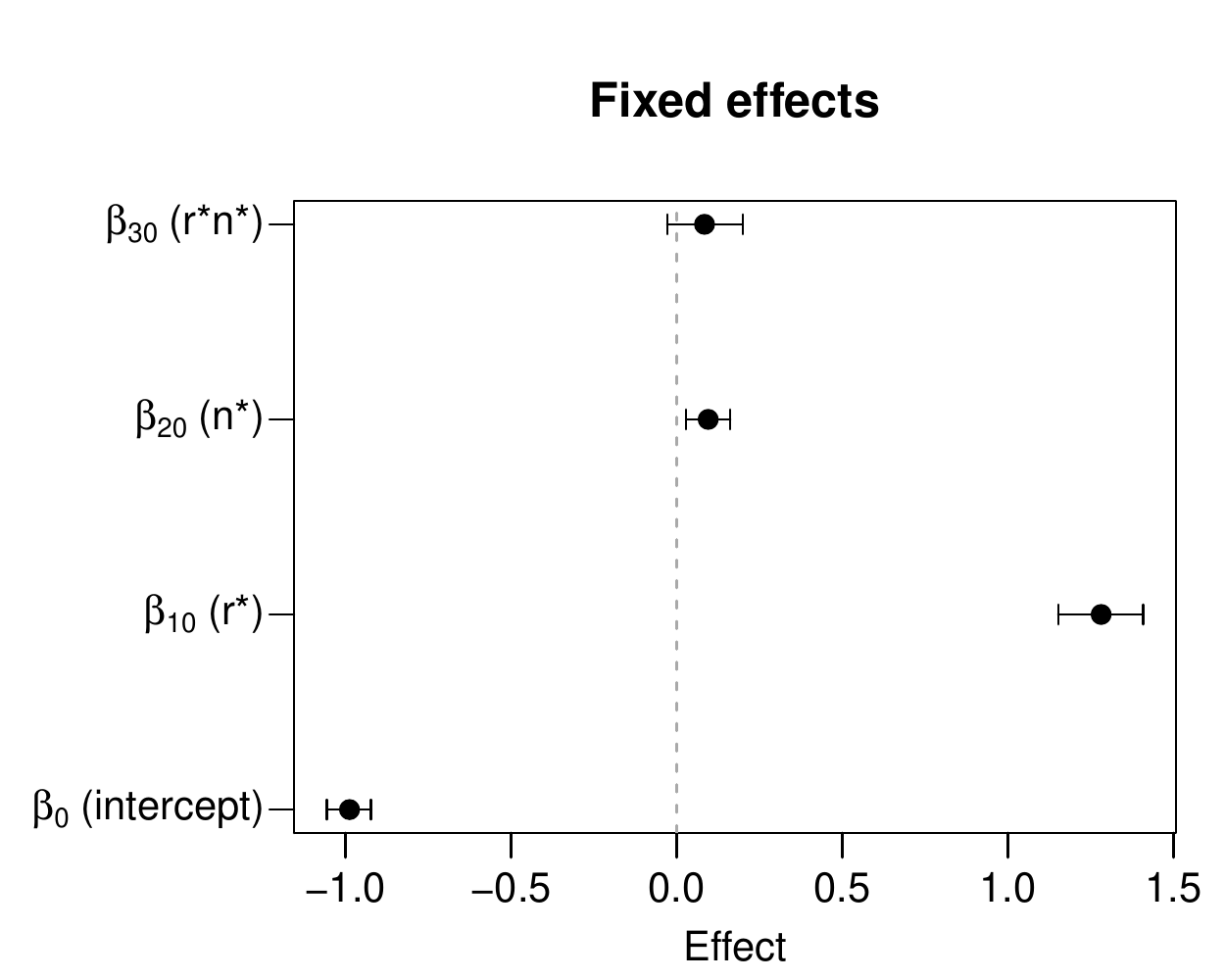}
  \caption{Fixed effects and bootstrap 95\% confidence intervals estimated for the second analysis model, depicting the overall intercept ($\beta_0$), the slope of the network size ($\beta_{10}$), the slope of the number of sources ($\beta_{20}$), and their interaction ($\beta_{30}$).}
  \label{fig:effects2}
\end{figure}

To gain better insight into differences across representation strategies, we used a second hierarchical model where the representation strategy was modeled as an ordinal variable $r^*$ instead of the nominal variable $r$ used in the first model. In particular, $r^*$ represents the size of the network, so we coded \emph{SS-R} as $0$, \emph{MS-SR@FC} as $0.2$, \emph{MS-CR@6} as $0.4$, \emph{MS-CR@4} as $0.6$, \emph{MS-CR@2} as $0.8$, and \emph{MSS-CR} as $1$ (see Fig.~\ref{fig:split}). In detail, this second model is as follows:
\begin{align}
y^*_i &= \beta_{0} +
	\beta_{1d[i]}\cdot r^*_i +
	\beta_{2d[i]}\cdot n^*_i +
	\beta_{3d[i]}\cdot r^*_i\cdot n^*_i +
	e_i &e_i&\sim N(0,\sigma^2_e) \label{eq:m21}\\
\beta_{1d} &= \beta_{10} + u_{1d} &u_{1d}&\sim N(0,\sigma^2_1) \label{eq:m22}\\
\beta_{2d} &= \beta_{20} + u_{2d} &u_{2d}&\sim N(0,\sigma^2_2) \label{eq:m23}\\
\beta_{3d} &= \beta_{30} + u_{3d} &u_{3d}&\sim N(0,\sigma^2_3) \label{eq:m24}.
\end{align}
In contrast to the first model, there is no representation-specific fixed intercept but an overall intercept $\beta_0$. The effect of the network size is similarly modeled as the sum of an overall fixed slope $\beta_{10}$ and a random dataset-specific effect $u_{1d}$. Likewise, this model includes the main effect of the number of sources (fixed effect $\beta_{20}$), as well as its interaction with the network size (fixed effect $\beta_{30}$). Fig.~\ref{fig:effects2} shows the fitted coefficients, confirming the statistically positive effect of the size of the networks and, to a smaller degree but still significant, of the number of sources. The interaction term is not statistically significant, probably because of the unclear benefit of the number of sources in \emph{MS-SR@FC}.

Overall, these analyses confirm that all multi-source strategies outperform the single-source representations, with a direct relation to the number of parameters in the network. In addition, there is a clearly positive effect of the number of sources, with a minor interaction between both factors. 

Fig.~\ref{fig:performance_by_n} also suggests that the variability of performance scores decreases with the number of learning sources used. This implies that if there are more learning sources available, one can expect less variability across instantiations of the network. Most importantly, variability obtained for a single learning source ($n=1$) is always larger than the variability with 2 or more sources. The Ballroom dataset shows much smaller variability when BPM is included in the combination. For this specific dataset, this indicates that once \emph{bpm} is used to learn the representation, the expected performance is stable and does not vary much, even if we keep including more sources. Section~\ref{res:single_vs_multi} provides more insight in this regard.

\subsection{Single-Source vs. Multi-Source}
\label{res:single_vs_multi}

\begin{figure}
  \centering\includegraphics [scale=\figscale]{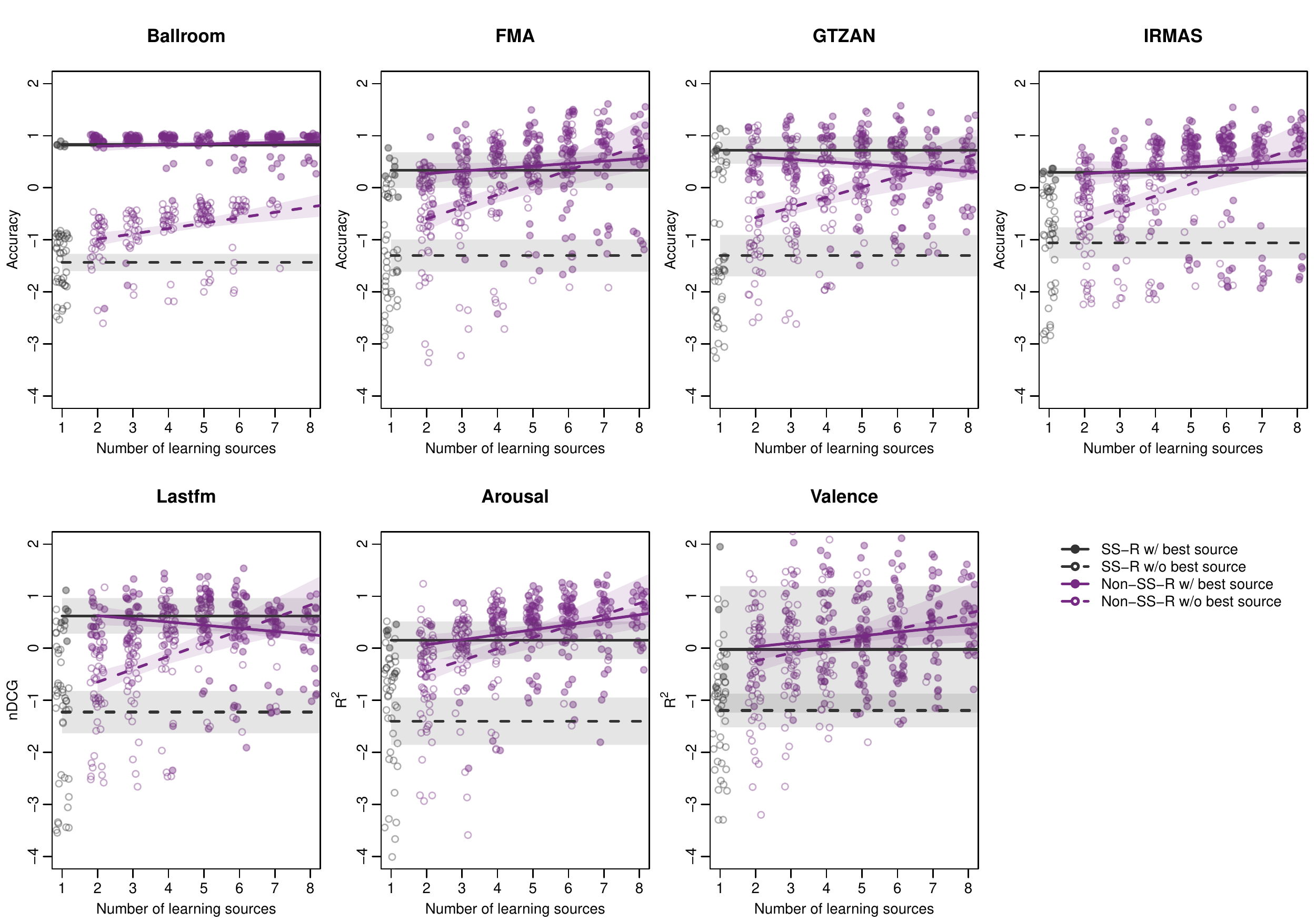}
  \caption{(Standardized) performance by number of learning sources. Solid points mark representations including the source performing best with \emph{SS-R} in the dataset; empty points mark representations without it. Solid and dashed lines represent linear fits, respectively; dashed areas represent 95\% confidence intervals.}
  \label{fig:performance_w_wo_best_stl}
\end{figure}

\begin{figure}
  \centering\includegraphics[scale=.4]{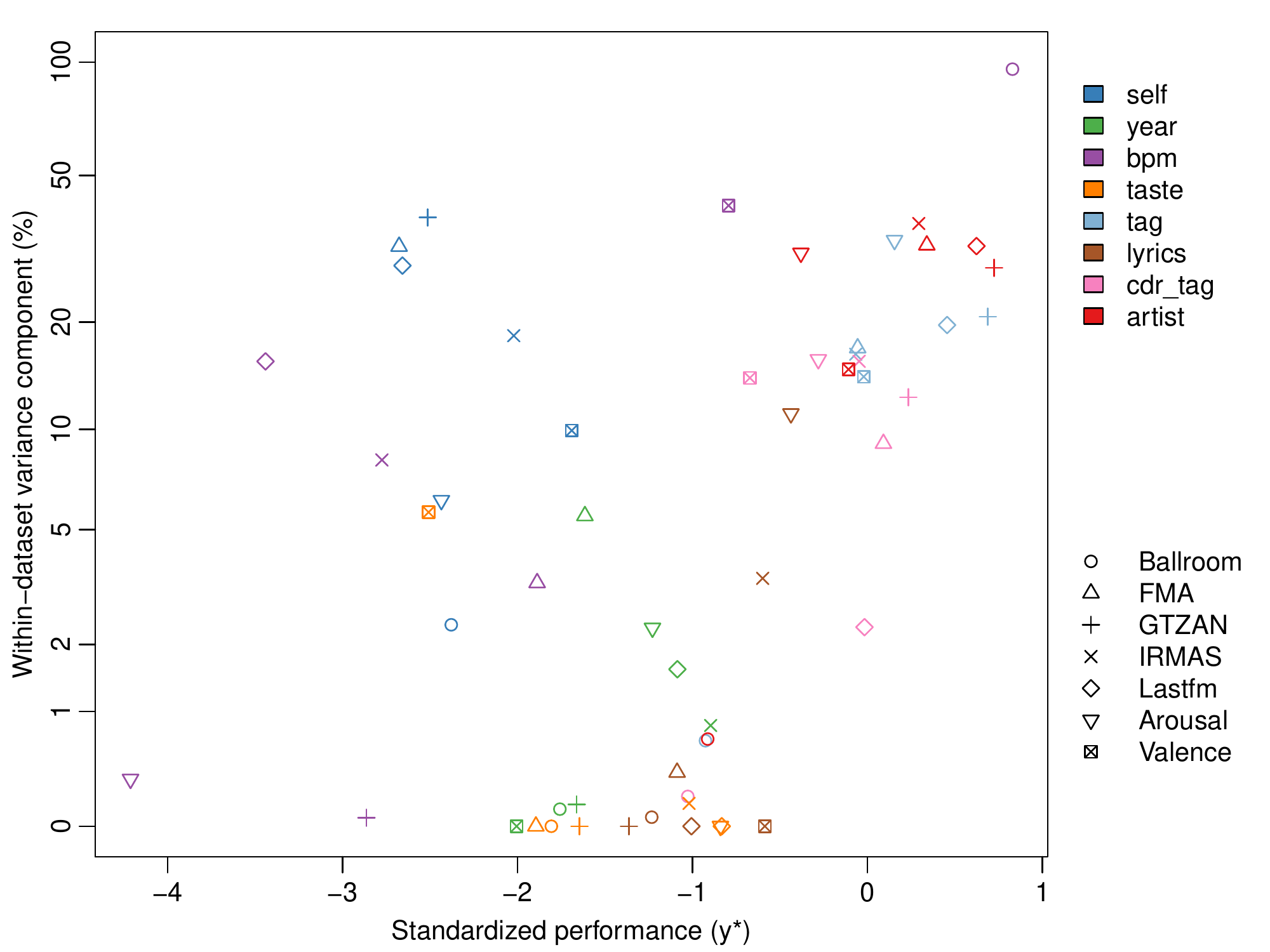}
  \caption{Correlation between (standardized) \emph{SS-R} performance and variance component.}
  \label{fig:rank_cor}
\end{figure}

The evidence so far tells us that, \emph{on average}, learning from multiple sources leads to better performance than learning from a single source. However, it could be possible that the \emph{SS-R} representation with the best learning source for the given target dataset still performs better than a multi-source alternative. In fact, in Fig.~\ref{fig:performance_by_n} there are many cases where the best \emph{SS-R} representation (black circles at $n=1$) already perform quite well compared to the more sophisticated alternatives.
Fig.~\ref{fig:performance_w_wo_best_stl} presents similar scatter plots, but now explicitly differentiating between representations using the single best source (filled circles, solid lines) and not using it (empty circles, dashed lines). The results suggest that even if the strongest learning source for the specific dataset is not used, the others largely compensate for it in the multi-source representations, catching up and even surpassing the best \emph{SS-R} representations. The exception to this rule is again \emph{bpm} in the Ballroom dataset, where it definitely makes a difference. As the plots shows, the variability for low numbers of learning sources is larger when not using the strongest source, but as more sources are added, this variability reduces.

To further investigate this issue, for each target dataset, we also computed the variance component due to each of the learning sources, excluding \emph{SS-R} representations~\cite{Searle2006variance}. A large variance due to one of the sources means that, on average and for that specific dataset, there is a large difference in performance between having that source or not. Table~\ref{tab:var} shows all variance components, highlighting the per-dataset largest. Apart from \emph{bpm} in the Ballroom dataset, there is no clear evidence that one single source is specially good in all datasets, which suggests that in general there is not a single source that one would use by default.
Notably though, sources \emph{artist}, \emph{tag} and \emph{self} tend to have large variance components.

\begin{table}[ht]
\centering
\caption{Variance components (as percent of total) of the learning sources, within each of the target datasets, and for non-\emph{SS-R} representations. Largest per dataset in bold face.}\label{tab:var}
\begin{tabular}{rrrrrrrr} \hline
 & Ballroom & FMA & GTZAN & IRMAS & Lastfm & Arousal & Valence \\ \hline
\emph{self} & 2 & \textbf{32} & \textbf{39} & 18 & 29 & 6 & 10 \\
\emph{year} & $<$1 & 6 & $<$1 & 1 & 2 & 2 & $<$1 \\
\emph{bpm} & \textbf{96} & 3 & $<$1 & 8 & 16 & $<$1 & \textbf{42} \\
\emph{taste} & $<$1 & $<$1 & $<$1 & $<$1 & $<$1 & $<$1 & 6 \\
\emph{tag} & 1 & 17 & 21 & 16 & 20 & \textbf{33} & 14 \\
\emph{lyrics} & $<$1 & $<$1 & $<$1 & 3 & $<$1 & 11 & $<$1 \\
\emph{cdr\_tag} & $<$1 & 9 & 12 & 16 & 2 & 16 & 14 \\
\emph{artist} & 1 & \textbf{32} & 28 & \textbf{37} & \textbf{32} & 31 & 15 \\ \hline
\end{tabular}
\end{table}

In addition, we observe that the sources with largest variance are not necessarily the sources that obtain the best results by themselves in an \emph{SS-R} representation (see Fig.~\ref{fig:single_task_rep}). We examined this relationship further by calculating the correlation between variance components and (standardized) performance of the corresponding \emph{SS-R}s. The Pearson correlation is $0.38$, meaning that there is a mild association. Fig.~\ref{fig:rank_cor} further shows this with a scatterplot, with a clear distinction between poorly-performing sources (\emph{year}, \emph{taste} and \emph{lyrics} at the bottom) and well-performing sources (\emph{tag}, \emph{cdr\_tag} and \emph{artist} at the right).

This result implies that even if some \emph{SS-R} is particularly strong for a given dataset, when considering more complex fusion architectures, the presence of that one source is not necessarily required because the other sources make up for its absence. This is especially important in practical terms, because different tasks generally have different best sources, and practitioners rarely have sufficient domain knowledge to select them up front. Also, and unlike the Ballroom dataset, many real-world problems are not easily solved with a single feature. Therefore, choosing a more general representation based on multiple sources is a much simpler way to proceed, which still yields comparable or better results.

In other words, if ``a single deep representation to rule them all'' is pre-trained, it is advisable to base this representation on multiple learning sources. At the same time, given that \emph{MSS-CR} representations also generally show strong performance (albeit that they will bring high dimensionality), and that they will come `for free' as soon as \emph{SS-R} networks are trained, alternatively, we could imagine an ecosystem in which the community could pre-train and release many \emph{SS-R} networks for different individual sources in a distributed way, and practitioners can then collect these into \emph{MSS-CR} representations, without the need for retraining.\par

\subsection{Compactness}
\label{res:task_num:compactness}

\begin{figure}
  \centering\includegraphics[width=0.7\textwidth]{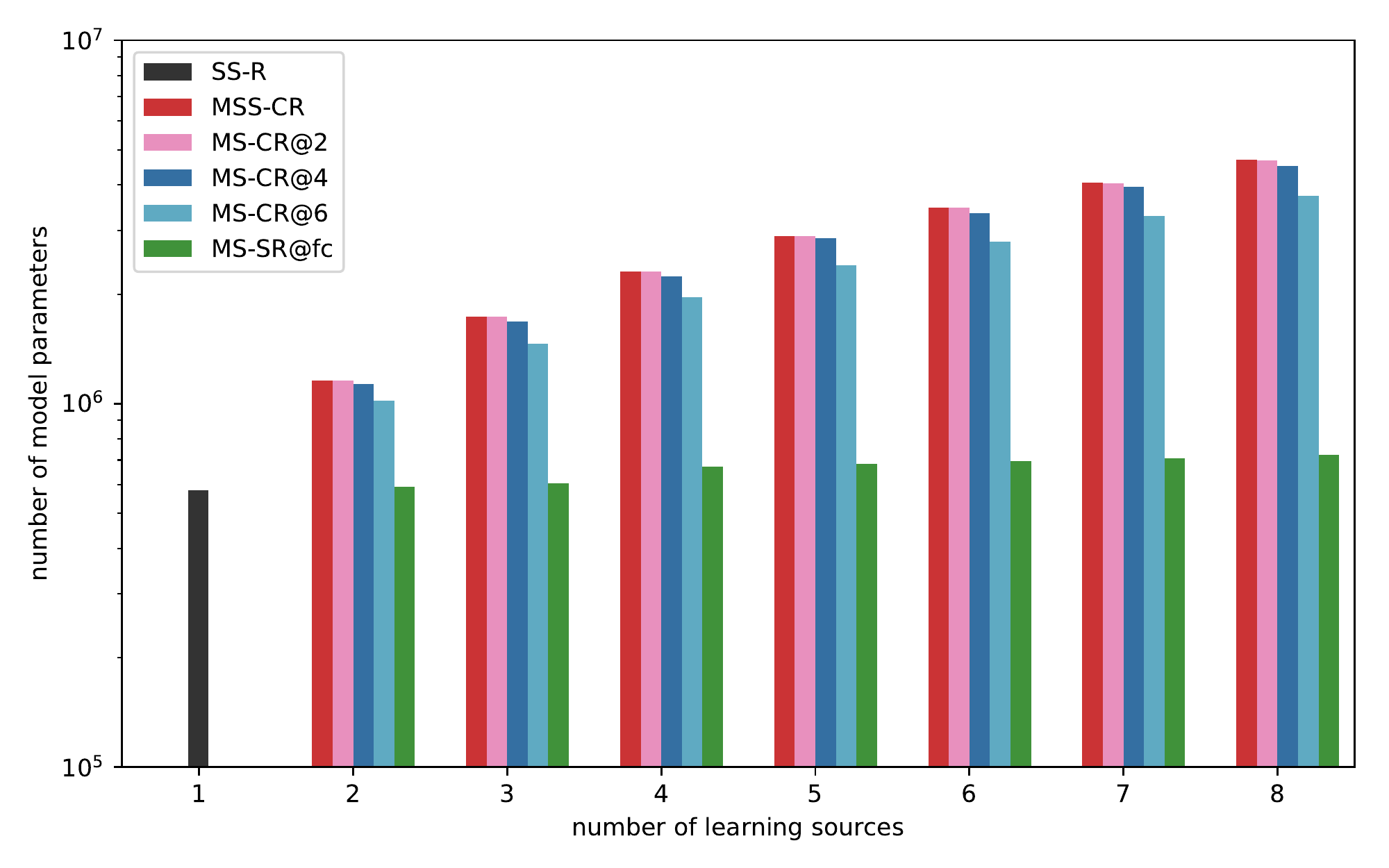}
  \caption{Number of network parameters by number of learning sources.}
  \label{fig:complexity}
\end{figure}

% \julian{review}
Under an MTDTL setup with branching (the \emph{MS-CR} architectures), as more learning sources are used, not only the representation will grow larger, but so will the necessary deep network to learn it: see Fig.~\ref{fig:complexity} for an overview of necessary model parameters for the different architectures. When using all the learning sources, \emph{MS-CR@6}, which for a considerable part encompasses a shared network architecture and branches out relatively late, has an around 6.3 times larger network size compared to the network size needed for \emph{SS-R}. In contrast, \emph{MS-SR@FC}, which is the most heavily shared MTDTL case, uses a network that is only 1.2 times larger than the network needed for \emph{SS-R}.\par

Also, while the representations resulting from the \emph{MSS-CR} and various \emph{MS-CR} architectures linearly depend on the chosen number of learning sources $m$ (see Table~\ref{tab:fusion}), for \emph{MS-SR@FC}, which has a fixed dimensionality of $d$ independent of $m$, we do notice increasing performance as more learning sources are used, except \emph{IRMAS} dataset. This implies that under MTDTL setups, the network does learn as much as possible from the multiple sources, even in case of fixed network capacity.\par

\subsection{Multiple Explanatory Factors}
\label{res:mulexpfac}

\begin{figure}
\centering
\includegraphics[width=0.95\textwidth]{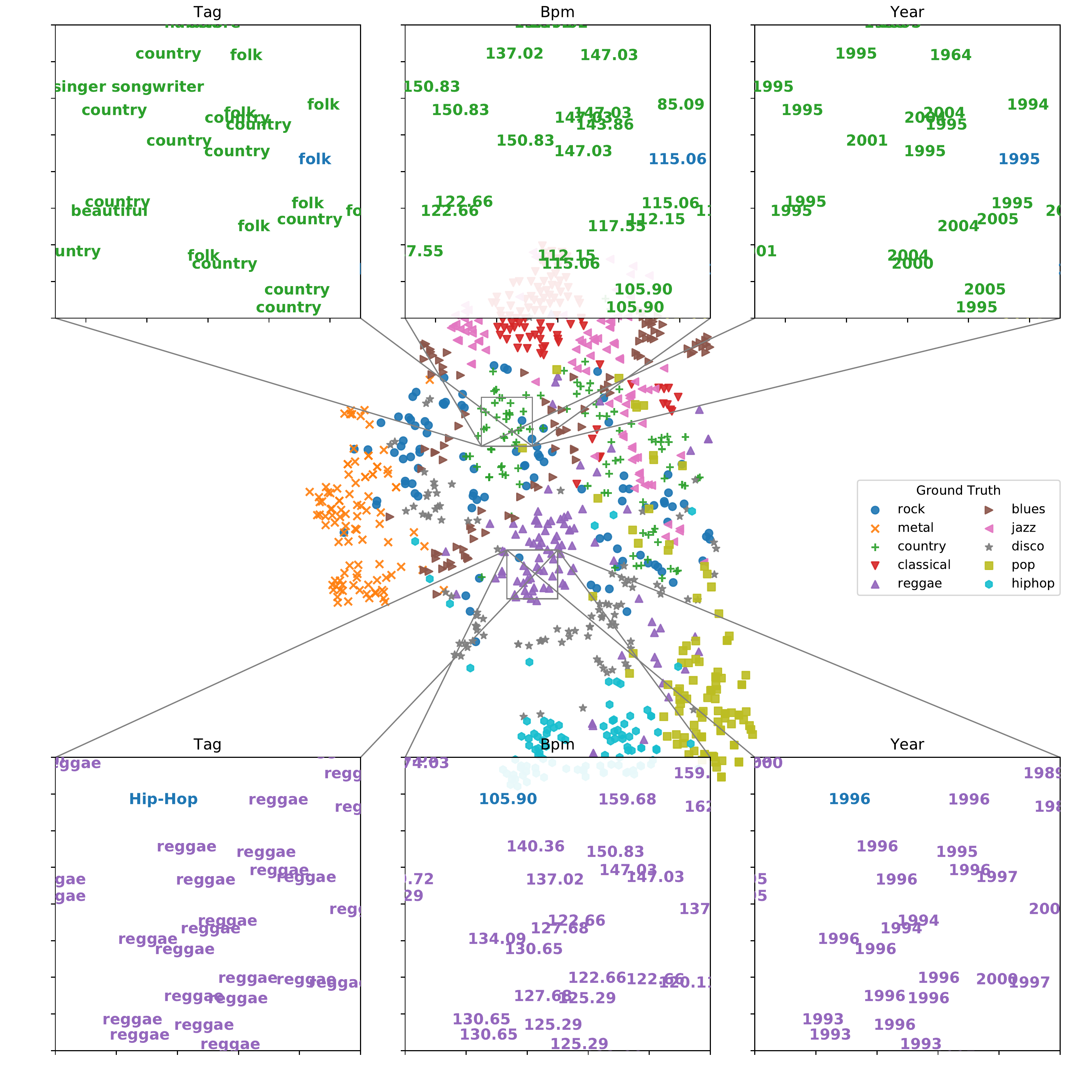}
\caption[The LOF Caption]{Potential semantic explainability of DTMTL music representations. Here, we provide a visualization using t-SNE~\cite{VanDerMaaten2008VisualizingT-sne}, plotting 2-dimensional coordinates of each sample from the GTZAN dataset, as resulting from an \emph{MS-CR} representation trained on 5 sources\footnotemark. In the zoomed-in panes, we overlay the strongest topic model terms in $z_t$, for various types of learning sources.}
\label{fig:topic_semantic}
\end{figure}
\footnotetext{The specific model used in the visualization is the $232$th model from the experimental design we introduce in Section~\ref{eval:expdesign}, which is performing better than 95\% of other models on GTZAN target dataset.}

By training representation models on multiple learning sources in the way we did, our hope is that the representation will reflect latent semantic facets that will ultimately allow for semantic explainability. In Fig.~\ref{fig:topic_semantic}, we show a visualization that suggests this indeed may be possible. More specifically, we consider one of our \emph{MS-CR} models trained on 5 learning sources. For each learning source-specific block of the representation, using the learning source-specific \texttt{fc-out} layers, we can predict a factor distribution $z_t$ for each of the learning sources. Then, from the predicted $z_t$, one can either map this back on the original learning labels $y_t$, or simply consider the strongest predicted topics (which we visualized in Fig.~\ref{fig:topic_semantic}), to relate the representation to human-understandable facets or descriptions.\footnote{Note that, as soon as a pre-trained representation network model will be adapted to an new dataset through transfer learning, the \texttt{fc-out} layer cannot be used to obtain such explanations from the learning sources used in the representation learning, since the layers will then be fine-tuned to another dataset. However, we hypothesize it may be possible that the semantic explainability can still be preserved, if fine-tuning is jointly conducted with the original learning sources used during the pre-training time in the multi-objective strategy.}\par

\section{Conclusion}
\label{concl}

In this paper, we have investigated the effect of different strategies to learn music representations with deep networks, considering multiple learning sources and different network architectures with varying degrees of shared information. Our main research questions are how the number and combination of learning sources (\textbf{RQ1}), and different configurations of the shared architecture (\textbf{RQ2}) affect effectiveness of the learned deep music representation. As a consequence, we conducted an experiment training 425 neural network models with different combinations of learning sources and architectures.

After an extensive empirical analysis, we can summarize our findings as follows:

\begin{itemize}
\item{\textbf{RQ1} The number of learning sources positively affects the effectiveness of a learned deep music representation, although representations based on a single learning source will already be effective in specialized cases (e.g. BPM and the Ballroom dataset).}
\item{\textbf{RQ2} In terms of architecture, the amount of shared information has a negative effect on performance: larger models with less shared information (e.g.~\emph{MS-CR@2}, \emph{MSS-CR}) tend to outperform models where sharing is higher (e.g.~\emph{MS-CR@6}, \emph{MS-SR@FC}), all of which outperform the base model (\emph{SS-R}).}
\end{itemize}
\par

Our findings give various pointers to useful future work. First of all, `generality' is difficult to define in the music domain, maybe more so than in CV or NLP, in which lower-level information atoms may be less multifaceted in nature (e.g.\ lower-level representations of visual objects naturally extend to many vision tasks, while an equivalent in music is harder to pinpoint). In case of clear task-specific data skews, practitioners should be pragmatic about this.

Also, we only investigated one special case of transfer learning, which might not be generalized well if one considers the adaptation of the pre-trained network for further fine-tuning with respect to their target dataset. Since there are various choices to make, which will bring substantial amount of variability, we decided to leave the aspects for further future works. We believe open-sourcing the models we trained throughout this work will be helpful for such follow-up works. Another limitation of current work is the selective set of label types in the learning sources. For instance, there are also a number of MIR related tasks that are using time-variant labels such as automatic music transcription, segmentation, beat tracking and chord estimation. We believe that such tasks should be investigated as well in the future to build a more complete overview of MTDTL problem.

Finally, in our current work, we still largely considered MTDTL as a `black box' operation, trying to learn \emph{how} MTDTL can be effective. However, the original reason for starting this work was not only to yield an effective general-purpose representation, but one that also would be semantically interpretable according to different semantic facets. We showed some early evidence our representation networks may be capable of picking up such facets; however, considerable future work will be needed into more in-depth analysis techniques of \emph{what} the deep representations actually learned.

\begin{acknowledgements}
This work was carried out on the Dutch national e-infrastructure with the support of SURF Cooperative. We further thank the CDR for having provided their album-level genre annotations for our experiments. We thank Keunwoo Choi for the discussion and all the help regarding the implementation of his work. We also thank David Tax for the valuable inputs and discussion. Finally, we thank editors and reviewers for their effort and constructive help to improve this work.
\end{acknowledgements}

\textit{Conflict of interest: Jaehun Kim, Juli\'{a}n Urbano, Cynthia C.~S. Liem and Alan Hanjalic state that there are no conflicts of interest.}

% \bibliographystyle{spmpsci}      % mathematics and physical sciences
% \bibliographystyle{plain}
\bibliographystyle{unsrtnat}
\bibliography{main}   % name your BibTeX data base

\end{document}